\documentclass{article} 
\usepackage{iclr2026_conference,times}

\usepackage{amsmath,amsfonts,bm}

\def\eqref#1{equation~\ref{#1}}

\def\1{\bm{1}}

\DeclareMathAlphabet{\mathsfit}{\encodingdefault}{\sfdefault}{m}{sl}
\SetMathAlphabet{\mathsfit}{bold}{\encodingdefault}{\sfdefault}{bx}{n}

\usepackage{url}

\usepackage[utf8]{inputenc} 
\usepackage[T1]{fontenc}    
\usepackage{url}            
\usepackage{booktabs}       
\usepackage{amsfonts}       
\usepackage{nicefrac}       
\usepackage{microtype}      
\usepackage{xcolor}         
\usepackage{natbib}
\usepackage[utf8]{inputenc} 
\usepackage[T1]{fontenc}    
\usepackage{url}            
\usepackage{booktabs}       
\usepackage{amsfonts}       
\usepackage{nicefrac}       
\usepackage{microtype}      
\usepackage[table]{xcolor}         
\usepackage{graphicx}
\usepackage{caption}
\usepackage{booktabs}
\usepackage{amsmath}
\usepackage{amssymb}
\usepackage{pifont}
\usepackage{subcaption}
\usepackage{stfloats}
\usepackage{wrapfig}
\usepackage{multirow}
\usepackage{amssymb}
\usepackage{xcolor}
\usepackage{xspace}
\usepackage{dsfont}
\usepackage{enumitem}
\usepackage{wrapfig}  
\usepackage{xcolor}

\definecolor{HyperlinkBlue}{RGB}{0, 102, 204} 
\definecolor{MyLinkColor}{HTML}{092997}  
\usepackage[colorlinks=true,urlcolor=HyperlinkBlue]{hyperref}
 
\usepackage{arydshln}
\usepackage{tcolorbox}
\tcbuselibrary{skins, breakable, theorems}

\definecolor{myred}{HTML}{F67280}
\definecolor{myblue}{HTML}{31ACD0}
\definecolor{mygreen}{HTML}{E0F9E0}
\definecolor{mypink}{HTML}{FFE8E8}
\hypersetup{
    colorlinks=true,
    linkcolor=myred,     
    citecolor=myblue,    
    urlcolor=myblue      
}

\usepackage{listings}
\usepackage{array}
\usepackage{scalerel}
\usepackage{soul}
\usepackage{color}
\usepackage{longtable}
\newcommand{\git}{\raisebox{-1.5pt}{\includegraphics[height=1.05em]{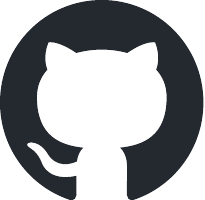}}\xspace}
\newcommand{\hf}{\raisebox{-1.5pt}{\includegraphics[height=1.05em]{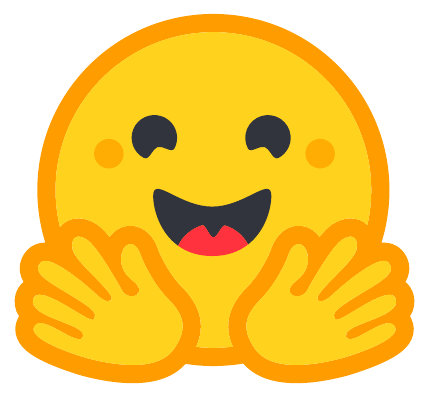}}\xspace}
\newcommand{\omni}{\raisebox{-1.5pt}{\includegraphics[height=1.05em]{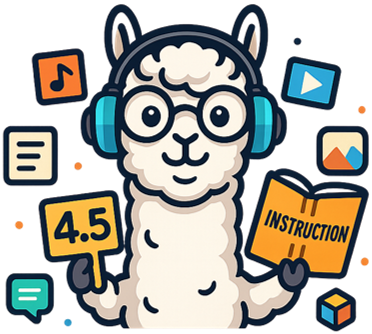}}\xspace}

\title{\includegraphics[scale=0.1]{fig/logo.png} Omni-Reward: Towards Generalist Omni-Modal Reward Modeling with Free-Form Preferences}

\author{Zhuoran Jin\textsuperscript{1,2,*}, Hongbang Yuan\textsuperscript{1,2,*}, Kejian Zhu\textsuperscript{1,2,*}, \\
\textbf{Jiachun Li\textsuperscript{1,2},} \textbf{Pengfei Cao\textsuperscript{1,2},} \textbf{Yubo Chen\textsuperscript{1,2},}  \textbf{Kang Liu\textsuperscript{1,2},} \textbf{Jun Zhao\textsuperscript{1,2}} \\ \textsuperscript{1}School of Artificial Intelligence, University of Chinese Academy of Sciences \\ \textsuperscript{2}Institute of Automation, Chinese Academy of Sciences\\
\texttt{\{zhuoran.jin, hongbang.yuan\} @nlpr.ia.ac.cn } \texttt{zhukejian2025@ia.ac.cn }\\
\texttt{\{jiachun.li, pengfei.cao, yubo.chen, kliu, jzhao\} @nlpr.ia.ac.cn } \\
}

\iclrfinalcopy
\begin{document}

\maketitle

\begin{abstract}
\vspace{-10pt}

Reward models (RMs) play a critical role in aligning AI behaviors with human preferences, yet they face two fundamental challenges: (1) \textbf{Modality Imbalance}, where most RMs are mainly focused on text and image modalities, offering limited support for video, audio, and other modalities; and
(2) \textbf{Preference Rigidity}, where training on fixed binary preference pairs fails to capture the complexity and diversity of personalized preferences.
To address the above challenges, we propose \texttt{Omni-Reward}, a step toward generalist omni-modal reward modeling with support for free-form preferences, consisting of: (1) \textbf{Evaluation}: We introduce \texttt{Omni-RewardBench}, the first omni-modal RM benchmark with free-form preferences, covering nine tasks across five modalities including text, image, video, audio, and 3D;
(2) \textbf{Data}: We construct \texttt{Omni-RewardData}, a multimodal preference dataset comprising 248K general preference pairs and 69K instruction-tuning pairs for training generalist omni-modal RMs;
(3) \textbf{Model}: We propose \texttt{Omni-RewardModel}, which includes both discriminative and generative RMs, and achieves strong performance on \texttt{Omni-RewardBench} as well as other widely used reward modeling benchmarks.

{\footnotesize 
\urlstyle{rm}
\begin{center}
    \renewcommand{\arraystretch}{1.2}
    \vspace{-4pt}
    \begin{tabular}{rcl}
         \hf & \textbf{Benchmark} & \url{https://hf.co/datasets/HongbangYuan/OmniRewardBench}\\
         \hf & \textbf{Dataset} & \url{https://hf.co/datasets/jinzhuoran/OmniRewardData}\\
         \hf & \textbf{Model} & \url{https://hf.co/jinzhuoran/OmniRewardModel} \\
          \git & \textbf{Code} & \url{https://github.com/HongbangYuan/OmniReward}\\
    \end{tabular}
\end{center}
}

\end{abstract}

\vspace{-5pt}
\section{Introduction}

\vspace{-5pt}

\def\thefootnote{}\footnotetext{* These authors contributed equally to this work.}\def\thefootnote{\arabic{footnote}}

To achieve more human-like intelligence \citep{shams2008benefits}, artificial general intelligence (AGI) is increasingly advancing toward an \textbf{omni-modal} paradigm \citep{nextgpt, llamaomni, showo}, where AI models are expected to process and generate information across diverse modalities (\textit{i.e.}, \textit{any-to-any} models).
Benefiting from the rapid progress in large language models (LLMs) \citep{ llama3, qwen25}, researchers are extending their powerful \textit{text-centric} capabilities to other modalities such as \textit{images}, \textit{video}, and \textit{audio}, enabling models (\textit{e.g.}, GPT‑4o \citep{openai2024gpt4o}, Gemini 2.0 Flash \citep{deepmind2025geminiflash}, and Qwen2.5-Omni \citep{qwenomni}) to not only understand multimodal inputs but also generate outputs using the most appropriate modality.

Despite the remarkable progress that existing omni-modal models have achieved on textual, visual, and auditory tasks, aligning their behaviors with human preferences remains a fundamental challenge \citep{alignanything,rlhfv, mmrlhf}. 
For example, models may fail to follow user instructions in speech-based interactions (\textit{i.e.}, \textit{helpfulness}), respond to sensitive prompts with harmful videos (\textit{i.e.}, \textit{harmlessness}), or generate hallucinated content when describing images (\textit{i.e.}, \textit{trustworthy}).
Reinforcement learning from human feedback (RLHF) \citep{DBLP:journals/corr/abs-1909-08593, instructgpt} has emerged as a promising approach for aligning model behaviors with human preferences.
RLHF integrates human feedback into the training loop by using it to guide the model toward more desirable and human-aligned responses.
This process \citep{rlhfworkflow} involves collecting human preference data to train a reward model (RM), which is subsequently used to fine-tune the original model through reinforcement learning by providing reward signals that guide its behavior.
Therefore, RMs play a pivotal role in RLHF, acting as a learned proxy of human preferences.

However, current RMs face two challenging problems: (1) \textbf{Modality Imbalance}: Most existing RMs \citep{ DBLP:journals/corr/abs-2407-06551, skywork, InternLMReward} predominantly focus on text and image modalities, while offering limited support for other modalities such as video and audio.
With the development of omni-modal models, achieving alignment in both understanding and generation across underrepresented modalities is becoming critically important;
(2) \textbf{Preference Rigidity}: 
Current preference data \citep{pick, skywork} is typically collected based on broadly accepted high-level values, such as helpfulness and harmlessness. RMs are then trained on these binary preference pairs, resulting in a fixed and implicit notion of preference embedded within the model.
Nevertheless, because human preferences cannot be neatly categorized into binary divisions, this paradigm fails to capture the diversity of personalized preferences \citep{DBLP:conf/nips/LeePKS24}.

Considering the above challenges, we propose \omni\texttt{Omni-Reward}, a step towards universal omni-modal reward modeling with free-form preferences.
For \textbf{modality imbalance}, \texttt{Omni-Reward} should be able to handle all modalities used in omni-modal models, including those that are rarely covered in existing preference data, such as video and audio.
It should also support reward shaping for complex multimodal tasks, such as image editing, video understanding, and audio generation, enabling a broad range of real-world applications.
For \textbf{preference rigidity}, \texttt{Omni-Reward} should not only capture general preferences grounded in widely shared human values, but also be capable of dynamically adjusting reward scores based on specific free-form preferences and multi-dimensional evaluation criteria.
To achieve this goal, we design \texttt{Omni-Reward} based on three key aspects:

\textbf{Evaluation}: RM evaluations \citep{rewardbench, rmbench, RMB} have primarily focused on text-only tasks, with recent efforts extending to visual understanding and generation \citep{hpdv2, vlrewardbench, mjbench}. Moreover, most RM benchmarks emphasize general preference judgments, while largely overlooking user-specific preferences and modality-dependent evaluation needs.
To address these gaps, we introduce \texttt{Omni-RewardBench}, an omni-modal reward modeling benchmark with free-form preferences, designed to evaluate the performance of RMs across diverse modalities.
Specifically, we collect prompts from various tasks and domains, elicit modality-specific responses from multiple models, and employ three annotators to provide free-form preference descriptions and label each response pair as \textit{chosen}, \textit{rejected}, or \textit{tied}.
Ultimately, \texttt{Omni-RewardBench} includes \textbf{3,725} high-quality human-annotated preference pairs, encompassing 9 distinct tasks and covering modalities such as text, image, video, audio, and 3D data.

\textbf{Data}: Current RMs are built upon large amounts of high-quality preference data. 
However, these preference datasets are typically designed for specific tasks and preferences, making it challenging for RMs to adapt to unseen multimodal tasks or  user preferences.
To enhance generalization, we construct \texttt{Omni-RewardData}, a large-scale multimodal preference dataset that spans a wide range of tasks.
We collect existing preference datasets to support general preference learning, and propose in-house instruction-tuning data to help RMs understand user preferences expressed in free-form language.
\texttt{Omni-RewardData} comprises \textbf{248K} general and \textbf{69K} fine-grained preference pairs.

\textbf{Model}: Building on \texttt{Omni-RewardData}, we further introduce two omni-modal reward models: \texttt{Omni-RewardModel-BT} and \texttt{Omni-RewardModel-R1}.
First, we train a discriminative RM named \texttt{Omni-RewardModel-BT} on the full \texttt{Omni-RewardData} using a classic Bradley–Terry objective.
Despite strong performance, its scoring process lacks transparency.
To address this, we explore a reinforcement learning approach to train a generative RM, named \texttt{Omni-RewardModel-R1}. It encourages the RM to engage in explicit reasoning by generating a textual critic in addition to producing a scalar score, and it is trained with only 3\% of the \texttt{Omni-RewardData}.

Built upon \texttt{Omni-RewardBench}, we conduct a thorough evaluation of multimodal large language models (MLLMs) used as generative RMs, including GPT-4o \citep{openai2024gpt4o}, Gemini-2.0 \citep{deepmind2025geminiflash}, Qwen2.5-VL \citep{qwen25vl}, and Gemma-3 \citep{gemma_2025}, as well as several purpose-built RMs for multimodal tasks, such as IXC-2.5-Reward \citep{IXC} and UnifiedReward \citep{wang2025unified}.
Our experimental results reveal the following findings:
(1) \texttt{Omni-RewardBench} presents significant challenges for current MLLMs, especially under the \textit{w/ Ties} setting.
The strongest commercial model, Claude 3.5 Sonnet \citep{anthropic2024claude35}, achieves the highest accuracy at \textbf{66.54\%}, followed closely by the open-source Gemma-3 27B at \textbf{65.12\%}, while existing purpose-built multimodal RMs still lag behind, indicating substantial room for improvement.
(2) There indeed exists the \textbf{modality imbalance} problem, particularly evident in the poor performance of existing models on tasks such as text-to-audio, text-to-3D, and text-image-to-image.
(3) RM performance is significantly correlated across various multimodal understanding (or generation) tasks, suggesting a certain degree of generalization potential within similar task categories.

Building on the findings above, we further evaluate how well \texttt{Omni-RewardModel} addresses the limitations of existing RMs.
Our experiments uncover the key insights below: (1) \texttt{Omni-RewardModel} achieves strong performance on \texttt{Omni-RewardBench}, attaining \textbf{73.68\%} accuracy under the \textit{w/o Ties} setting and \textbf{65.36\%} accuracy under the \textit{w/ Ties} setting, and shows strong generalization to challenging tasks.
(2) \texttt{Omni-RewardModel} also captures general human preferences and achieves performance comparable to or even better than the state-of-the-art (SOTA) on public RM benchmarks such as VL-RewardBench \citep{vlrewardbench} and Multimodal RewardBench \citep{MultimodalRewardBench}.
(3) Instruction-tuning is crucial for RMs, as it effectively alleviates the \textbf{preference rigidity} issue and enables the model to dynamically adjust reward scores according to free-form user preferences.
In summary, our contributions are as follows:

(1) We present \texttt{Omni-RewardBench}, the first omni-modal reward modeling benchmark with free-form preferences, designed to systematically evaluate the performance of RMs across diverse modalities. 
It includes nine multimodal tasks and 3,725 high-quality preference pairs, posing significant challenges to existing multimodal RMs, revealing substantial room for improvement.

(2) We construct \texttt{Omni-RewardData}, a multimodal preference dataset comprising 248K general preference pairs and 69K newly collected instruction-tuning pairs with free-form preference descriptions, enabling RMs to generalize across modalities and align with diverse user preferences.

(3) We propose \texttt{Omni-RewardModel}, including the discriminative  \texttt{Omni-RewardModel-BT} and the generative  \texttt{Omni-RewardModel-R1}.
Our model not only demonstrates significant improvement on \texttt{Omni-RewardBench}, with a \textbf{20\%} accuracy gain over the base model, but also achieves performance comparable to or even exceeding that of SOTA RMs on public benchmarks.

\vspace{-10pt}

\section{Omni-RewardBench}
\vspace{-5pt}

In this section, we introduce \texttt{Omni-RewardBench}, an omni-modal reward modeling benchmark with free-form preferences for evaluating the RM performance across diverse modalities.
Table \ref{dataset_comparison} presents a comprehensive comparison between \texttt{Omni-RewardBench} and existing multimodal reward modeling benchmarks.
\texttt{Omni-RewardBench} covers 9 tasks across image, video, audio, text, and 3D modalities, and incorporates free-form preferences to support evaluating RMs under diverse criteria.
Figure~\ref{dataset_construction_workflow} illustrates the overall construction workflow, including prompt collection (\textsection~\ref{dataset_collection}), response generation (\textsection~\ref{dataset_collection}), criteria annotation (\textsection~\ref{dataset_annotation}), and preference annotation (\textsection~\ref{dataset_annotation}).

\vspace{-5pt}

\subsection{Task Definition and Setting}
\label{task_definition}

Each data sample in \texttt{Omni-RewardBench} is represented as $(x, y_1, y_2, c, p)$, where $x$ denotes the input prompt, $y_1$ and $y_2$ are two candidate responses generated by AI models, $c$ specifies the free-form user preference or evaluation criterion, and $p$ indicates the preferred response under the given criterion $c$.
An effective RM is expected to correctly predict $p$ given $(x, y_1, y_2, c)$.
We provide two evaluation settings:
(1) \textit{w/o Ties} (ties-excluded), where $p \in \{y_1, y_2\}$, requiring a strict preference between the two responses;
(2) \textit{w/ Ties} (ties-included), a more challenging setting where $p \in \{y_1, y_2, \text{tie}\}$, allowing for the case where the two responses are equally preferred under the given criterion.

\vspace{-5pt}

\subsection{Dataset Collection}
\label{dataset_collection}

\begin{figure}[t!]
    \centering
    \includegraphics[width=0.86\linewidth]{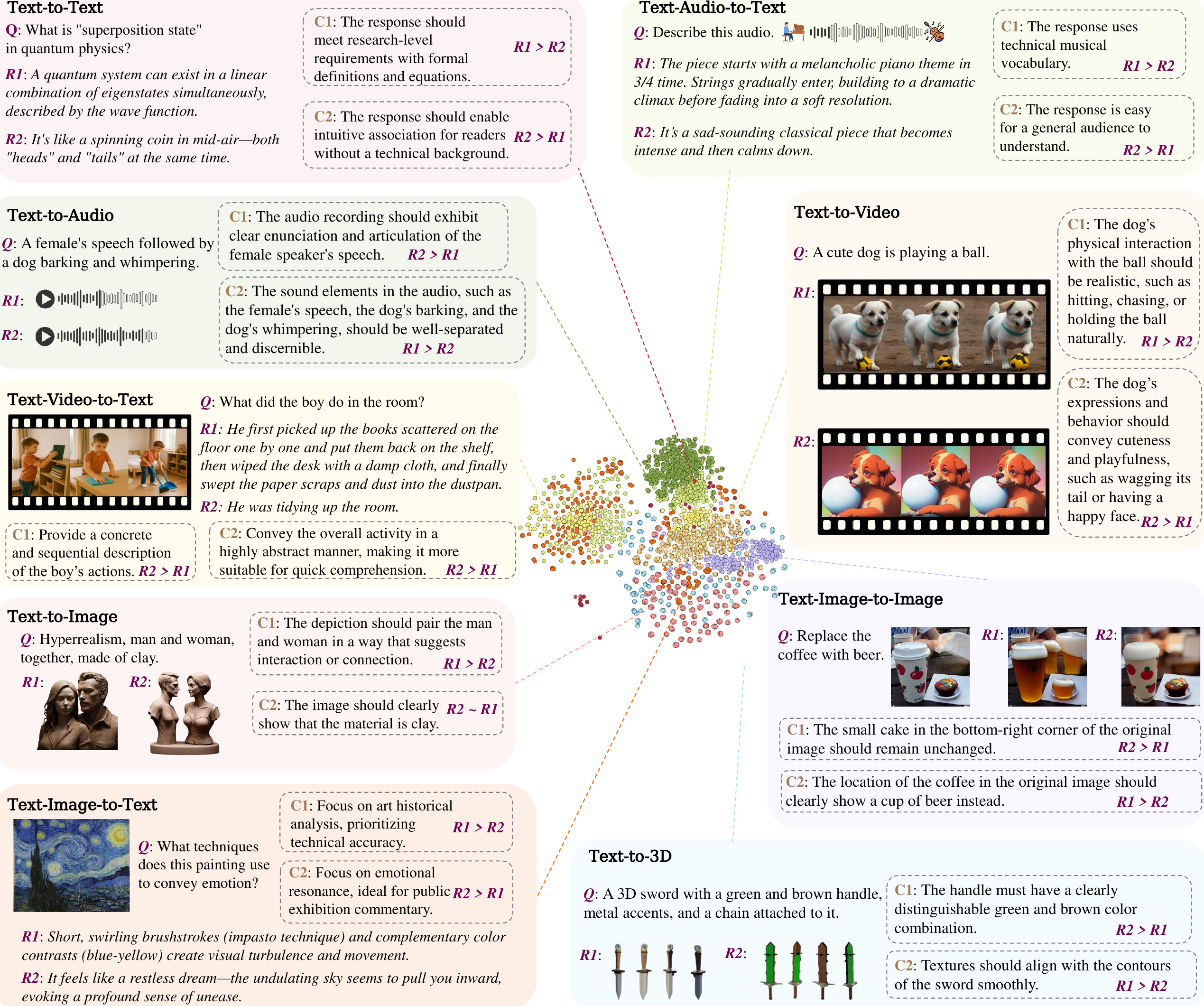}
    \caption{Illustration of nine reward modeling tasks in \texttt{Omni-RewardBench}.}
    \label{benchmark_description}
    \vspace{-20pt}
\end{figure}

Figure~\ref{benchmark_description} provides an overview of the nine tasks covered in \texttt{Omni-RewardBench}, spanning a wide range of modalities. Detailed descriptions of each task are provided below.

\textbf{Text-to-Text (T2T)}: T2T refers to the text generation task of outputting textual responses based on user instructions, which represents a fundamental capability of LLMs. 
In this task, $x$ denotes the user instruction, and $y$ denotes the textual response. We collect prompts from real-world downstream tasks across diverse scenarios in RMB \citep{RMB} and RPR \citep{DBLP:conf/nips/PitisXRS24}, covering tasks like open QA, coding, and reasoning.
Subsequently, we include responses generated by 13 LLMs.

\textbf{Text-Image-to-Text (TI2T)}: TI2T denotes the image understanding task of generating textual responses based on textual instructions and image inputs.
In this task, $x$ represents a pair consisting of a user instruction and an image, and $y$ denotes the textual response. We consider image understanding tasks with varying levels of complexity. We first collect general instructions from VL-Feedback \citep{li-etal-2024-vlfeedback}, and subsequently gather meticulously constructed, layered, and complex instructions from MIA-Bench \citep{qian2025miabenchbetterinstructionfollowing}. The responses are collected from 14 MLLMs.

\textbf{Text-Video-to-Text (TV2T)}: TV2T refers to the video understanding task of generating textual responses based on both textual instructions and video inputs. In this task, $x$ indicates a user instruction and a video, and $y$ indicates the corresponding textual response.
We collect video-question pairs from VCGBench-Diverse \citep{maaz2024videogptintegratingimagevideo}, which contains a range of video categories and diverse user questions. The durations of the selected videos range from 30 s to 358 s, with an average of 207 s. We collect responses from 4 MLLMs equipped with video understanding capabilities. 

\textbf{Text-Audio-to-Text (TA2T)}: TA2T denotes the audio understanding task of generating textual responses based on both textual instructions and audio inputs. In this task, $x$ denotes the paired input of a user instruction and an audio clip, and $y$ denotes the textual response. We collect diverse, open-ended questions from OpenAQA \citep{DBLP:conf/iclr/0001LLKG24}, each paired with an approximately 10 s audio clip. Subsequently, responses are collected from 4 MLLMs capable of audio understanding.

\textbf{Text-to-Image (T2I)}: T2I denotes the image synthesis task of generating high-fidelity images based on user textual prompts. In this task, $x$ denotes the  textual description, and $y$ denotes the generated image. We collect diverse manually-written prompts that reflect the general interests of model users, along with corresponding images from Rapidata \citep{rapidata2024humanstyle} and HPDv2 \citep{hpdv2}, covering 27 text-to-image models ranging from autoregressive-based to diffusion-based architectures.

\textbf{Text-to-Video (T2V)}: T2V denotes the video synthesis task of generating temporally coherent videos from textual descriptions. In this task, $x$ denotes the input textual description, and $y$ denotes the corresponding generated video. We collect human-written prompts from GenAI-Bench \citep{jiang2024genai} and subsequently acquire the corresponding videos generated by up to 8 text-to-video models.

\textbf{Text-to-Audio (T2A)}: T2A denotes the audio generation task of synthesizing audio clips with temporal and semantic consistency from textual descriptions. In this task, $x$ denotes the textual description, and $y$ denotes the generated audio. We collect various prompts from Audio-alpaca \citep{majumder2024tango} and responses from the latent diffusion model Tango \citep{ghosal2023texttoaudiogenerationusinginstructiontuned}.

\textbf{Text-to-3D (T23D)}: T23D denotes the 3D generation task of synthesizing three-dimensional objects from textual descriptions. In this task, $x$ is the  textual prompt, and $y$ denotes the generated 3D object. We collect user prompts from 3DRewardDB \citep{ye2024dreamreward} and responses from the multi-view diffusion model mvdream-sd2.1-diffusers \citep{DBLP:conf/iclr/ShiWYMLY24}. The responses are presented in the multi-view rendered format of each 3D object, enabling direct image-based input to MLLMs.

\textbf{Text-Image-to-Image (TI2I)}: TI2I denotes the image editing task of modifying an image based on textual instructions. In this task, $x$ denotes a source image and an editing prompt, and $y$ denotes the edited image. We collect images to be edited and user editing prompts from GenAI-Bench \citep{jiang2024genai}. The responses are generated with a broad range of diffusion models.

\subsection{Criteria and Preference Annotation}
\label{dataset_annotation}

Following the collection of user prompts and corresponding responses, the evaluation criteria $c$ and the user preference $p$ are subsequently annotated.
For the criteria annotation, each annotator manually creates multiple evaluation criteria in textual form based on the input $x$.
For the preference annotation, each data sample is independently labeled by three annotators based on the free-form evaluation criteria.
To ensure data quality, we first discarded 23\% of instances with invalid criteria annotations, followed by 15\% with conflicting preferences.
The entire annotation process is conducted by three PhD students in computer science, guided by detailed guidelines and supported by an annotation platform in Appendix \ref{Annotation Details}. Ethics and quality control during data annotation are detailed in Appendix \ref{echics_and_quality_control}.
A total of 3,725 preference data are finally collected, covering 9 tasks across all modalities. More detailed statistics of \texttt{Omni-RewardBench} are provided in Table \ref{appendix:dataset_statistics} and Table \ref{tab:number_of_criteria_per_pair}.

\vspace{-10pt}

\section{Omni-RewardModel}

\vspace{-5pt}

In this section, we first construct \texttt{Omni-RewardData}, a multimodal preference dataset comprising 248K general preference pairs and 69K newly collected instruction-tuning pairs with free-form preference descriptions for RM training.
Based on the dataset, we propose two omni-modal RMs: \texttt{Omni-RewardModel-BT} (discriminative RM) and \texttt{Omni-RewardModel-R1} (generative RM).

\vspace{-5pt}

\subsection{Omni-RewardData Construction}

\begin{wraptable}{r}{0.5\textwidth}
\vspace{-16pt}
\caption{Data statistics of \texttt{Omni-RewardData}. * denotes the subset constructed in this work.}
\centering
\small
\begin{tabular}{llc}
\toprule
\textbf{Task} & \textbf{Subset} & \textbf{\#Size} \\
\midrule
\multirow{3}{*}{T2T} & Skywork-Reward-Preference & 50,000 \\
                     & Omni-Skywork-Reward-Preference* & 16,376 \\
                     & Omni-UltraFeedback* & 7,901 \\
\midrule
\multirow{4}{*}{T2I} & HPDv2 & 50,000 \\
                     & EvalMuse & 2,944 \\
                     & Omni-HPDv2* & 8,959 \\
                     & Omni-Open-Image-Preferences* & 8,105 \\
\midrule
\multirow{4}{*}{TI2T} & RLAIF-V & 83,124 \\
                      & OmniAlign-V-DPO & 50,000 \\
                      & Omni-RLAIF-V* & 15,867 \\
                      & Omni-VLFeedback* & 12,311 \\
\midrule
\multirow{2}{*}{T2V} & VideoDPO & 10,000 \\
                     & VisionRewardDB-Video & 1,795 \\
\bottomrule
\end{tabular}
\vspace{-12pt}
\label{tab:omni-data}
\end{wraptable}

High-quality and diverse human preference data is crucial for training effective omni-modal RMs.
However, existing preference datasets are often limited in scope because they focus on specific tasks or general preferences.
This limitation hinders the model's ability to generalize to novel multimodal scenarios and adapt to multiple user preferences.
To improve the generalization ability of RMs, we construct \texttt{Omni-RewardData}, which primarily covers four task types: T2T, TI2T, T2I, and T2V, and comprises a total of 317K preference pairs, including both general and fine-grained preferences.

Specifically, we first collect a substantial amount of existing preference datasets to help the model learn general preferences. The details are as follows:
(1) For \textbf{T2T}, we select 50K data from Skywork-Reward-Preference \citep{skywork}, a high-quality dataset that provides binary preference pairs covering a wide range of instruction-following tasks.
(2) For \textbf{TI2T}, we use select 83K data from RLAIF-V \citep{rlaifv}, a multimodal preference dataset that targets trustworthy alignment and hallucination reduction of MLLMs. Moreover, we also include 50K data from OmniAlign-V-DPO \citep{OmniAlign-V}, which features diverse images, open-ended questions, and varied response formats.
(3) For \textbf{T2I}, we sample 50K data from HPDv2 \citep{hpdv2}, a well-annotated dataset containing human preference judgments on images generated by text-to-image generative models. In addition, we adopt EvalMuse \citep{evalmuse}, which provides large-scale human annotations covering both overall and fine-grained aspects of image-text alignment.
(4) For \textbf{T2V}, we collect 10K samples from VideoDPO \citep{VideoDPO}, which evaluates both the visual quality and semantic alignment.
We also integrate 2K preference pairs from VisionReward \citep{VisionReward}.

Moreover, as these data primarily reflect broadly accepted and general preferences, RMs trained solely on them often struggle to adapt reward assignment based on user-specified fine-grained preferences or customized evaluation criteria.
Therefore, we propose constructing instruction-tuning data specifically for RMs, where each data instance is formatted as $(c,x,y_1,y_2,p)$.
We first sample preference pairs $(x, y_1, y_2)$ from existing datasets, and prompt GPT-4o to generate a free-form instruction $c$ reflecting a user preference that supports either $y_1$ or $y_2$, together with the corresponding label $p$. To ensure quality, we use GPT-4o-mini, Qwen2.5-VL 7B, and Gemma-3-12B-it to verify the consistency of $(c, x, y_1, y_2)$ with the label $p$.
We obtain the following in-house subset: (1) For \textbf{T2T}, we construct 24K data based on Skywork-Reward-Preference \citep{skywork} and UltraFeedback \citep{ULTRAFEEDBACK}.
(2) For \textbf{TI2T}, we synthesize 28K data based on RLAIF-V and VLFeedback \citep{li-etal-2024-vlfeedback}.
(3) For \textbf{T2I}, we generate 17K data using HPDv2 and Open-Image-Preferences \citep{open-image-preferences-v1}.
The statistics of \texttt{Omni-RewardData} are shown in Table \ref{tab:omni-data}.

\vspace{-5pt}

\subsection{Discriminative Reward Modeling with Bradley-Terry}

\begin{figure}[t]
    \centering
    \includegraphics[width=\linewidth]{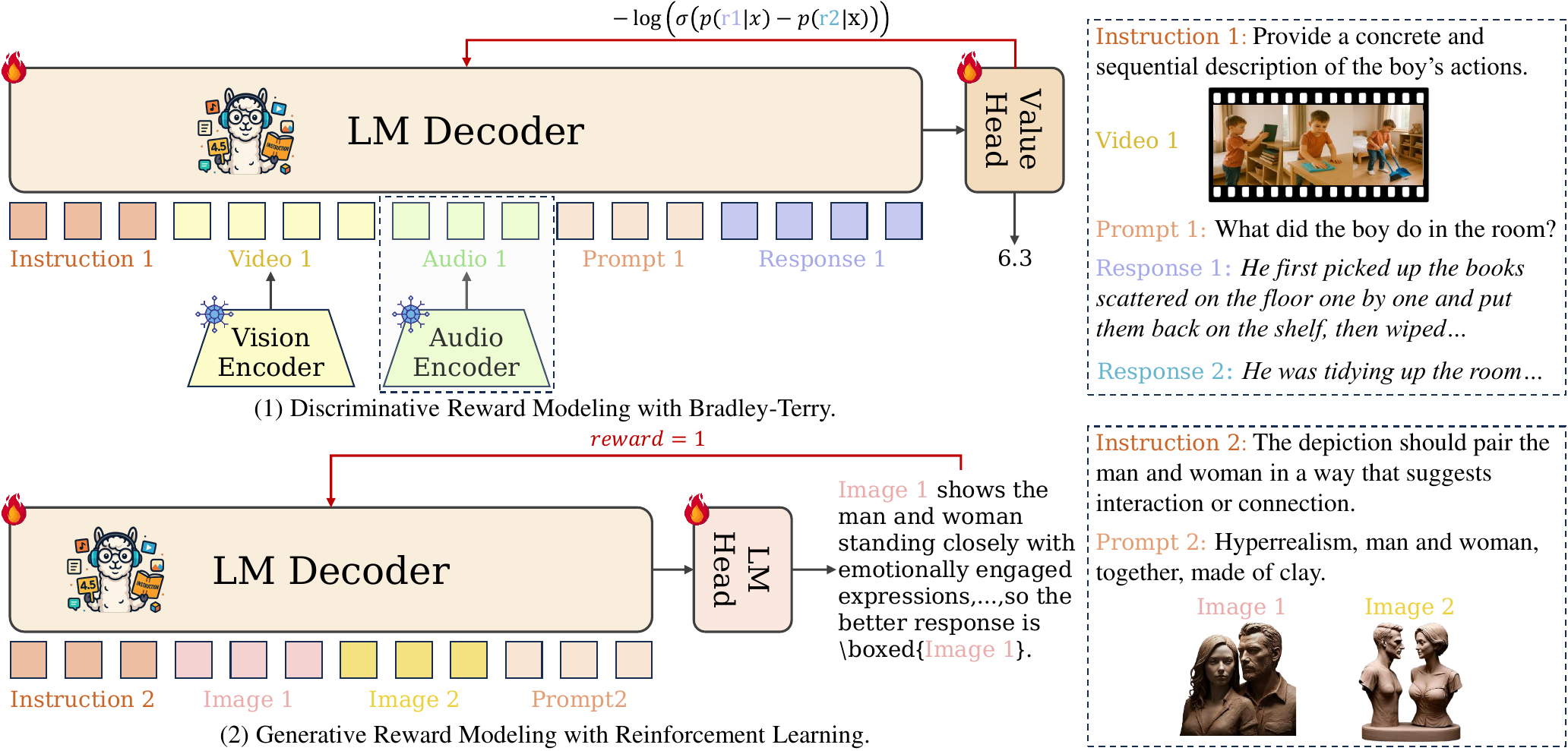}
    \caption{Overview of the architecture of \texttt{Omni-RewardModel}.}
    \label{model}
        \vspace{-15pt}
\end{figure}

Following standard practice in reward modeling, we adopt the Bradley-Terry loss \citep{bradley1952rank} for training our discriminative RM where a scalar score is assigned to each candidate response:
\begin{equation}
\mathcal{L}_{\text{BT}} = -\log \frac{\exp(r_{\text{BT}}(c, x, y_c))}{\exp(r_{\text{BT}}(c, x, y_c)) + \exp(r_{\text{BT}}(c, x, y_r))},
\end{equation}
where $c$ denotes an optional instruction that specifies user preference, $y_c$ denotes the chosen response, $y_r$ denotes the rejected response, $r_{\text{BT}}(\cdot)$ denotes the reward function.
Specifically, we train \texttt{Omni-RewardModel-BT} on \texttt{Omni-RewardData} using MiniCPM-o-2.6 \citep{minicpmv}.
As shown in Figure \ref{model}(1), we freeze the parameters of the vision and audio encoders, and only update the language model decoder and the value head.
User-specific preferences and task-specific evaluation criteria are provided as system messages, allowing the RM to adapt its scoring behavior accordingly.

\vspace{-5pt}

\subsection{Generative Reward Modeling with Reinforcement Learning}
\vspace{-5pt}

To improve the interpretability of the reward scoring process, we further explore a reinforcement learning approach for training a pairwise generative reward model, denoted as \texttt{Omni-RewardModel-R1}. 
As shown in Figure \ref{model}(2), given the input $(c, x, y_1, y_2)$, the model $r_{\text{R1}}(\cdot)$ is required to first generate a Chain-of-Thought (CoT) explanation $e$, followed by a preference prediction $p'$.
We optimize the model using the GRPO-based reinforcement learning \citep{deepseekr1}, where the reward signal is computed by comparing the predicted preference $p'$ with the ground-truth preference $p$.
We train \texttt{Omni-RewardModel-R1} from scratch on 10K samples from \texttt{Omni-RewardData}, using Qwen2.5-VL-7B-Instruct \citep{qwen25vl} as the base model, without distillation from larger models.

\vspace{-10pt}

\section{Experiments}

\vspace{-5pt}

In this section, we conduct a comprehensive evaluation of a wide range of multimodal reward models, including generative RMs based on MLLMs and specialized RMs trained for task-specific objectives, as well as our proposed \texttt{Omni-RewardModel}.
Moreover, we also extend the evaluation to include widely adopted benchmarks from prior work in multimodal reward modeling.

\vspace{-5pt}

\subsection{Baseline Reward Models} 
\vspace{-5pt}

\textbf{Generative Reward Models.} We evaluate 30 generative RMs built upon state-of-the-art MLLMs, including 24 open-source and 6 proprietary models.
The open-source models cover both omni-modal (\textit{e.g.}, Phi-4 \citep{phi4}, Qwen2.5-Omni \citep{qwenomni}, MiniCPM-o-2.6 \citep{minicpmv}) and vision-language models (\textit{e.g.}, Qwen2-VL \citep{qwen2vl}, Qwen2.5-VL \citep{qwen25vl}, InternVL2.5 \citep{intern2_5}, InternVL3 \citep{zhu2025internvl3exploringadvancedtraining}, and Gemma3 \citep{gemma_2025}), with sizes ranging from 3B to 72B.
For proprietary models, we consider the GPT \citep{gpt4}, Gemini \citep{deepmind2025geminiflash}, and Claude \citep{anthropic2024claude3} series.
Specifically, we use GPT-4o-Audio-Preview in place of GPT-4o for the TA2T and T2A tasks.

\begin{table}[t]
\centering
\caption{Evaluation results on \texttt{Omni-RewardBench} under the \textit{w/ Tie} setting.}
\label{evaluation_result_w_tie}
 \resizebox{\linewidth}{!}{

\begin{tabular} 
{
>{\columncolor[HTML]{FDFAF6}}l 
>{\columncolor[HTML]{fbf4f5}}c  
>{\columncolor[HTML]{fef0e7}}c  
>{\columncolor[HTML]{fffef1}}c  
>{\columncolor[HTML]{f3f7ec}}c  
>{\columncolor[HTML]{fff5f5}}c  
>{\columncolor[HTML]{fffaf3}}c  
>{\columncolor[HTML]{f3f7ec}}c  
>{\columncolor[HTML]{f5fcfe}}c  
>{\columncolor[HTML]{f9f8ff}}c  
>{\columncolor[HTML]{CDF5FD}}c }

\toprule
\textbf{Model} & \textbf{T2T} & \textbf{TI2T} & \textbf{TV2T} & \textbf{TA2T} & \textbf{T2I} & \textbf{T2V} & \textbf{T2A} & \textbf{T23D} & \textbf{TI2I} & \textbf{Overall} \\
\midrule
\multicolumn{11}{c}{\textit{Open-Source Models}} \\ 
\href{https://huggingface.co/microsoft/Phi-4-multimodal-instruct}{Phi-4-Multimodal-Instruct} & 70.98 & 53.60 & 62.53 & 55.74 & 35.36 & 32.14 & 44.77 & 24.17 & 22.71 & 44.67  \\
\href{https://huggingface.co/Qwen/Qwen2.5-Omni-7B}{Qwen2.5-Omni-7B} & 65.71 & 55.11 & 56.66 & 59.66 & 55.99 & 50.85 & 32.60 & 43.71 & 43.23 & 51.50 \\
\href{https://huggingface.co/openbmb/MiniCPM-o-2\_6}{MiniCPM-o-2.6} & 61.39 & 51.89 & 60.95 & 60.50 & 47.35 & 39.70 & 21.90 & 37.09 & 39.30 & 46.67 \\
\href{https://huggingface.co/openbmb/MiniCPM-V-2\_6}{MiniCPM-V-2.6} & 57.55 & 54.73 & 53.27 & - & 48.92 & 44.61 & - & 39.40 & 36.68 & 47.88 \\
\href{https://huggingface.co/llava-hf/llava-onevision-qwen2-7b-ov-hf}{LLaVA-OneVision-7B-ov} & 50.84 & 42.23 & 45.37 & - & 43.42 & 40.08 & - & 35.43 & 37.12 & 42.07 \\
\href{https://huggingface.co/mistralai/Mistral-Small-3.1-24B-Instruct-2503}{Mistral-Small-3.1-24B-Instruct-2503} & 74.58 & 57.98 & 68.62 & - & 58.55 & 59.92 & - & 60.60 & 62.88 & 63.30 \\
\href{https://huggingface.co/Skywork/Skywork-R1V-38B}{Skywork-R1V-38B} & 77.94 & 59.47 & 67.72 & - & 47.94 & 45.94 & - & 43.71 & 41.92 & 54.95 \\
\href{https://huggingface.co/Qwen/Qwen2-VL-7B-Instruct}{Qwen2-VL-7B-Instruct} & 63.55 & 55.30 & 59.37 & - & 33.20 & 61.25 & - & 42.38 & 10.04 & 46.44 \\
\href{https://huggingface.co/Qwen/Qwen2.5-VL-3B-Instruct}{Qwen2.5-VL-3B-Instruct} & 53.00 & 49.05 & 51.24 & - & 47.74 & 51.23 & - & 45.36 & 44.54 & 48.88 \\
\href{https://huggingface.co/Qwen/Qwen2.5-VL-7B-Instruct}{Qwen2.5-VL-7B-Instruct} & 68.59 & 53.03 & 68.40 & - & 60.51 & 47.83 & - & 50.99 & 41.05 & 55.77 \\
\href{https://huggingface.co/Qwen/Qwen2.5-VL-32B-Instruct}{Qwen2.5-VL-32B-Instruct} & 74.82 & 60.23 & 63.88 & - & 60.51 & 62.38 & - & 62.58 & 69.43 & 64.83 \\
\href{https://huggingface.co/Qwen/Qwen2.5-VL-72B-Instruct}{Qwen2.5-VL-72B-Instruct} & 76.98 & 61.17 & 68.40 & - & 58.94 & 56.52 & - & 59.60 & 62.01 & 63.37 \\
\href{https://huggingface.co/OpenGVLab/InternVL2\_5-4B}{InternVL2\_5-4B} & 57.55 & 50.76 & 55.30 & - & 48.72 & 47.07 & - & 47.35 & 47.16 & 50.56 \\
\href{https://huggingface.co/OpenGVLab/InternVL2\_5-8B}{InternVL2\_5-8B} & 60.43 & 49.62 & 54.63 & - & 54.42 & 49.53 & - & 42.72 & 44.10 & 50.78 \\
\href{https://huggingface.co/OpenGVLab/InternVL2\_5-26B}{InternVL2\_5-26B} & 64.75 & 57.01 & 62.98 & - & 56.97 & 49.72 & - & 57.28 & 48.03 & 56.68 \\
\href{https://huggingface.co/OpenGVLab/InternVL2\_5-38B}{InternVL2\_5-38B} & 69.06 & 54.73 & 64.56 & - & 54.81 & 40.26 & - & 55.96 & 46.72 & 55.16 \\
\href{https://huggingface.co/OpenGVLab/InternVL2\_5-8B-MPO}{InternVL2\_5-8B-MPO} & 65.95 & 52.46 & 68.17 & - & 56.97 & 52.55 & - & 52.98 & 41.05 & 55.73 \\
\href{https://huggingface.co/OpenGVLab/InternVL2\_5-26B-MPO}{InternVL2\_5-26B-MPO} & 70.74 & 60.98 & \textbf{70.43} & - & 58.74 & 47.26 & - & 56.95 & 48.03 & 59.02 \\
\href{ https://huggingface.co/OpenGVLab/InternVL3-8B}{InternVL3-8B} & 76.02 & 58.71 & 67.95 & - & 57.37 & 48.77 & - & 51.66 & 43.67 & 57.74 \\
\href{https://huggingface.co/OpenGVLab/InternVL3-9B}{InternVL3-9B} & 73.86 & 57.39 & 66.59 & - & 57.37 & 51.80 & - & 60.93 & 47.16 & 59.30 \\
\href{https://huggingface.co/OpenGVLab/InternVL3-14B}{InternVL3-14B} & 76.74 & 61.74 & 68.62 & - & 60.51 & 61.25 & - & 59.27 & 55.02 & 63.31 \\
\href{https://huggingface.co/google/gemma-3-4b-it}{Gemma-3-4B-it} & 74.34 & 56.82 & 68.40 & - & 60.31 & 60.30 & - & 54.64 & 54.15 & 61.28 \\
\href{https://huggingface.co/google/gemma-3-12b-it}{Gemma-3-12B-it} & 73.62 & 58.52 & 66.14 & - & 59.33 & 62.57 & - & 56.95 & 56.33 & 61.92 \\
\href{https://huggingface.co/google/gemma-3-27b-it}{Gemma-3-27B-it} & 77.22 & 61.17 & 67.04 & - & 59.14 & 61.44 & - & 63.91 & 65.94 & 65.12 \\
\midrule
\multicolumn{11}{c}{\textit{Proprietary Models}} \\ 
\href{https://openai.com/index/hello-gpt-4o/}{GPT-4o} & \textbf{78.18} & 61.74 & 69.30 & 62.75 & 59.33 & 65.03 & 44.53 & \textbf{70.86} & \textbf{69.87} & 64.62 \\
\href{https://ai.google.dev/gemini-api/docs/models}{Gemini-1.5-Flash} & 72.90 & 58.52 & 68.62 & 57.42 & 62.48 & 63.52 & 32.85 & 62.25 & 63.32 & 60.21 \\
\href{https://ai.google.dev/gemini-api/docs/models}{Gemini-2.0-Flash} & 74.10 & 54.92 & 60.50 & 61.90 & 62.28 & 67.49 & 31.87 & 68.54 & 65.50 & 60.79 \\
\href{https://openai.com/index/gpt-4o-mini-advancing-cost-efficient-intelligence/}{GPT-4o-mini} & 76.50 & 60.23 & 67.95 & - & 57.56 & 65.22 & - & 60.26 & 60.26 & 64.00 \\
\href{https://docs.anthropic.com/en/docs/about-claude/models/all-models}{Claude-3-5-Sonnet-20241022} & 76.74 & 61.55 & 67.04 & - & 61.69 & 64.27 & - & 68.54 & 65.94 & \textbf{66.54} \\
\href{https://docs.anthropic.com/en/docs/about-claude/models/all-models}{Claude-3-7-Sonnet-20250219-Thinking} & 75.78 & \textbf{63.83} & 68.85 & - & 62.28 & 62.38 & - & 68.21 & 63.76 & 66.44 \\
\midrule
\multicolumn{11}{c}{\textit{Specialized Models}} \\ 
\href{https://huggingface.co/yuvalkirstain/PickScore\_v1}{PickScore} & 42.93 & 43.56 & 46.95 & - & 60.12 & 66.92 & - & 59.27 & 51.53 & 53.04 \\
\href{https://huggingface.co/xswu/HPSv2}{HPSv2} & 43.41 & 45.27 & 44.70 & - & \textbf{63.85} & 64.65 & - & 61.26 & 55.02 & 54.02 \\
\href{https://huggingface.co/internlm/internlm-xcomposer2d5-7b-reward}{InternLM-XComposer2.5-7B-Reward} & 59.95 & 52.65 & 65.69 & - & 45.19 & 61.25 & - & 43.05 & 9.61 & 48.20 \\
\href{https://huggingface.co/CodeGoat24/UnifiedReward-7b}{UnifiedReward} & 60.19 & 53.22 & 69.53 & - & 59.72 & \textbf{70.32} & - & 59.93 & 42.36 & 59.32 \\
\href{https://huggingface.co/CodeGoat24/UnifiedReward-7b-v1.5}{UnifiedReward1.5} & 59.47 & 54.17 & 69.30 & - & 58.35 & 69.57 & - & 61.59 & 45.41 & 59.69  \\

\hdashline
\texttt{Omni-RewardModel-R1}  & 71.22 & 56.06 & 63.88 & - & 61.69 & 58.22 & - & 63.91 & 46.29 & 60.18 \\
\texttt{Omni-RewardModel-BT}  & 75.30 & 60.23 & 68.85 & \textbf{70.59} & 58.35 & 64.08 & \textbf{63.99} & 67.88 & 58.95 & 65.36 \\

\midrule

Average & 67.32 & 55.52 & 63.02 & 59.66 & 55.31 & 55.59 & 34.75 & 53.98 & 48.60 & 56.68 \\
\bottomrule
\end{tabular}
}
\vspace{-15pt}
\end{table}

\begin{table}[h]
\centering
\caption{Evaluation results on \texttt{Omni-RewardBench} under the \textit{w/o Tie} setting.}
\label{evaluation_result_w_o_tie}
 \resizebox{\linewidth}{!}{

\begin{tabular} 
{
>{\columncolor[HTML]{FDFAF6}}l 
>{\columncolor[HTML]{fbf4f5}}c  
>{\columncolor[HTML]{fef0e7}}c  
>{\columncolor[HTML]{fffef1}}c  
>{\columncolor[HTML]{f3f7ec}}c  
>{\columncolor[HTML]{fff5f5}}c  
>{\columncolor[HTML]{fffaf3}}c  
>{\columncolor[HTML]{f3f7ec}}c  
>{\columncolor[HTML]{f5fcfe}}c  
>{\columncolor[HTML]{f9f8ff}}c  
>{\columncolor[HTML]{CDF5FD}}c }

\toprule
\textbf{Model} & \textbf{T2T} & \textbf{TI2T} & \textbf{TV2T} & \textbf{TA2T} & \textbf{T2I} & \textbf{T2V} & \textbf{T2A} & \textbf{T23D} & \textbf{TI2I} & \textbf{Overall} \\
\midrule
\multicolumn{11}{c}{\textit{Open-Source Models}} \\ 
\href{https://huggingface.co/microsoft/Phi-4-multimodal-instruct}{Phi-4-Multimodal-Instruct} & 81.15 & 68.14 & 74.74 & 63.47 & 46.03 & 51.72 & 55.05 & 39.02 & 49.28 & 58.73 \\
\href{https://huggingface.co/Qwen/Qwen2.5-Omni-7B}{Qwen2.5-Omni-7B} & 82.79 & 68.14 & 78.16 & 63.77 & 65.53 & 63.09 & 50.76 & 56.44 & 54.11 & 64.75 \\
\href{https://huggingface.co/openbmb/MiniCPM-o-2\_6}{MiniCPM-o-2.6} & 74.04 & 66.05 & 71.58 & 69.76 & 58.50 & 61.16 & 54.80 & 54.92 & 48.79 & 62.18 \\
\href{https://huggingface.co/openbmb/MiniCPM-V-2\_6}{MiniCPM-V-2.6} & 74.86 & 65.12 & 69.47 & - & 57.37 & 58.15 & - & 51.14 & 53.62 & 61.39 \\
\href{https://huggingface.co/llava-hf/llava-onevision-qwen2-7b-ov-hf}{LLaVA-OneVision-7B-ov} & 66.67 & 57.67 & 53.42 & - & 51.93 & 51.72 & - & 43.94 & 43.48 & 52.69 \\
\href{https://huggingface.co/mistralai/Mistral-Small-3.1-24B-Instruct-2503}{Mistral-Small-3.1-24B-Instruct-2503} & 84.43 & 65.79 & 79.47 & - & 65.99 & 68.67 & - & 67.80 & 71.98 & 72.02 \\
\href{https://huggingface.co/Skywork/Skywork-R1V-38B}{Skywork-R1V-38B} & \textbf{88.25} & 74.42 & 76.84 & - & 55.10 & 57.94 & - & 45.83 & 52.66 & 64.43 \\
\href{https://huggingface.co/Qwen/Qwen2-VL-7B-Instruct}{Qwen2-VL-7B-Instruct} & 79.78 & 70.00 & 76.58 & - & 37.41 & 68.03 & - & 47.35 & 12.08 & 55.89 \\
\href{https://huggingface.co/Qwen/Qwen2.5-VL-3B-Instruct}{Qwen2.5-VL-3B-Instruct} & 68.58 & 66.05 & 60.00 & - & 52.15 & 60.09 & - & 51.89 & 53.62 & 58.91 \\
\href{https://huggingface.co/Qwen/Qwen2.5-VL-7B-Instruct}{Qwen2.5-VL-7B-Instruct} & 80.87 & 66.28 & 78.95 & - & 65.53 & 64.59 & - & 64.77 & 50.72 & 67.39 \\
\href{https://huggingface.co/Qwen/Qwen2.5-VL-32B-Instruct}{Qwen2.5-VL-32B-Instruct} & 86.34 & 74.19 & 77.37 & - & 70.29 & 70.39 & - & 68.56 & 70.05 & 73.88 \\
\href{https://huggingface.co/Qwen/Qwen2.5-VL-72B-Instruct}{Qwen2.5-VL-72B-Instruct} & 87.70 & 74.65 & \textbf{80.53} & - & 71.88 & 67.17 & - & 66.67 & 69.57 & 74.02 \\
\href{https://huggingface.co/OpenGVLab/InternVL2\_5-4B}{InternVL2\_5-4B} & 69.95 & 63.49 & 64.47 & - & 58.50 & 54.94 & - & 50.38 & 41.55 & 57.61 \\
\href{https://huggingface.co/OpenGVLab/InternVL2\_5-8B}{InternVL2\_5-8B} & 72.13 & 64.88 & 65.00 & - & 64.40 & 61.59 & - & 58.33 & 53.14 & 62.78 \\
\href{https://huggingface.co/OpenGVLab/InternVL2\_5-26B}{InternVL2\_5-26B} & 77.60 & 72.79 & 76.32 & - & 68.03 & 62.88 & - & 68.56 & 59.90 & 69.44 \\
\href{https://huggingface.co/OpenGVLab/InternVL2\_5-38B}{InternVL2\_5-38B} & 84.15 & 66.05 & 70.53 & - & 66.67 & 63.30 & - & 68.94 & 57.97 & 68.23 \\
\href{https://huggingface.co/OpenGVLab/InternVL2\_5-8B-MPO}{InternVL2\_5-8B-MPO} & 75.96 & 65.12 & 77.63 & - & 65.99 & 61.80 & - & 62.88 & 55.07 & 66.35 \\
\href{https://huggingface.co/OpenGVLab/InternVL2\_5-26B-MPO}{InternVL2\_5-26B-MPO} & 80.87 & 73.72 & \textbf{80.53} & - & 68.93 & 62.66 & - & 67.80 & 60.87 & 70.77 \\
\href{ https://huggingface.co/OpenGVLab/InternVL3-8B}{InternVL3-8B} & 84.70 & 71.63 & 76.84 & - & 69.39 & 65.67 & - & 59.85 & 53.62 & 68.81 \\
\href{https://huggingface.co/OpenGVLab/InternVL3-9B}{InternVL3-9B} & 83.06 & 70.23 & 78.42 & - & 65.31 & 65.67 & - & 71.97 & 58.45 & 70.44 \\
\href{https://huggingface.co/OpenGVLab/InternVL3-14B}{InternVL3-14B} & 85.79 & 74.65 & 77.11 & - & 72.79 & 68.24 & - & 68.56 & 58.94 & 72.30 \\
\href{https://huggingface.co/google/gemma-3-4b-it}{Gemma-3-4B-it} & 83.88 & 73.02 & 77.37 & - & 72.34 & 66.09 & - & 67.05 & 63.77 & 71.93 \\
\href{https://huggingface.co/google/gemma-3-12b-it}{Gemma-3-12B-it} & 81.69 & 72.09 & 78.42 & - & 71.20 & 71.03 & - & 67.05 & 65.70 & 72.45 \\
\href{https://huggingface.co/google/gemma-3-27b-it}{Gemma-3-27B-it} & \textbf{88.25} & 75.58 & 78.16 & - & 68.48 & 71.03 & - & 73.86 & 71.50 & 75.27 \\

\midrule
\multicolumn{11}{c}{\textit{Proprietary Models}} \\ 

\href{https://openai.com/index/hello-gpt-4o/}{GPT-4o} & 86.89 & 75.58 & 77.11 & 70.96 & 69.61 & 73.18 & 53.28 & 77.65 & \textbf{73.91} & 73.13 \\
\href{https://ai.google.dev/gemini-api/docs/models}{Gemini-1.5-Flash} & 83.88 & 69.53 & 78.16 & 62.28 & 71.43 & 71.89 & 40.66 & 74.24 & 73.43 & 69.50 \\
\href{https://ai.google.dev/gemini-api/docs/models}{Gemini-2.0-Flash} & 85.25 & 67.91 & 75.26 & 67.96 & 70.52 & 74.25 & 60.86 & \textbf{79.17} & 71.98 & 72.57 \\
\href{https://openai.com/index/gpt-4o-mini-advancing-cost-efficient-intelligence/}{GPT-4o-mini} & 87.43 & 74.65 & 77.89 & - & 67.80 & 74.89 & - & 71.59 & 66.67 & 74.42 \\
\href{https://docs.anthropic.com/en/docs/about-claude/models/all-models}{Claude-3-5-Sonnet-20241022} & \textbf{88.25} & \textbf{76.28} & 78.68 & - & 70.75 & 72.53 & - & 77.65 & 72.46 & \textbf{76.66} \\
\href{https://docs.anthropic.com/en/docs/about-claude/models/all-models}{Claude-3-7-Sonnet-20250219-Thinking} & 84.43 & \textbf{76.28} & 77.89 & - & 70.07 & 70.60 & - & 76.89 & 72.46 & 75.52 \\

\midrule
\multicolumn{11}{c}{\textit{Specialized Models}} \\ 

\href{https://huggingface.co/yuvalkirstain/PickScore\_v1}{PickScore} & 49.18 & 53.49 & 54.47 & - & 69.61 & 75.97 & - & 67.05 & 57.49 & 61.04 \\
\href{https://huggingface.co/xswu/HPSv2}{HPSv2} & 49.18 & 55.12 & 51.58 & - & \textbf{73.70} & 73.61 & - & 70.45 & 60.87 & 62.07 \\
\href{https://huggingface.co/internlm/internlm-xcomposer2d5-7b-reward}{InternLM-XComposer2.5-7B-Reward} & 68.85 & 64.19 & 74.74 & - & 51.47 & 68.24 & - & 46.59 & 56.04 & 61.45 \\
\href{https://huggingface.co/CodeGoat24/UnifiedReward-7b}{UnifiedReward} & 68.58 & 59.77 & 79.47 & - & 68.93 & \textbf{79.83} & - & 68.56 & 46.86 & 67.43 \\
\href{https://huggingface.co/CodeGoat24/UnifiedReward-7b-v1.5}{UnifiedReward1.5} & 67.76 & 67.39 & 78.68 & - & 67.57 & 78.97 & - & 70.45 & 50.72 & 68.79 \\
\hdashline
\texttt{Omni-RewardModel-R1}  & 81.77 & 69.53 & 75.53 & - & 71.20 & 62.02 & - & 72.35 & 55.56 & 69.71 \\
\texttt{Omni-RewardModel-BT} & 85.79 & 72.79 & 79.47 & \textbf{75.45} & 67.12 & 72.75 & \textbf{66.41} & 77.65 & 65.70 & 73.68 \\
\midrule
Average & 78.38 & 68.57 & 73.77 & 66.37 & 64.61 & 66.62 & 52.57 & 63.54 & 58.10 & 67.29 \\
\bottomrule
\end{tabular}
}
\vspace{-15pt}
\end{table}

\begin{figure}[th]
    \centering
    \includegraphics[width=0.9\linewidth]{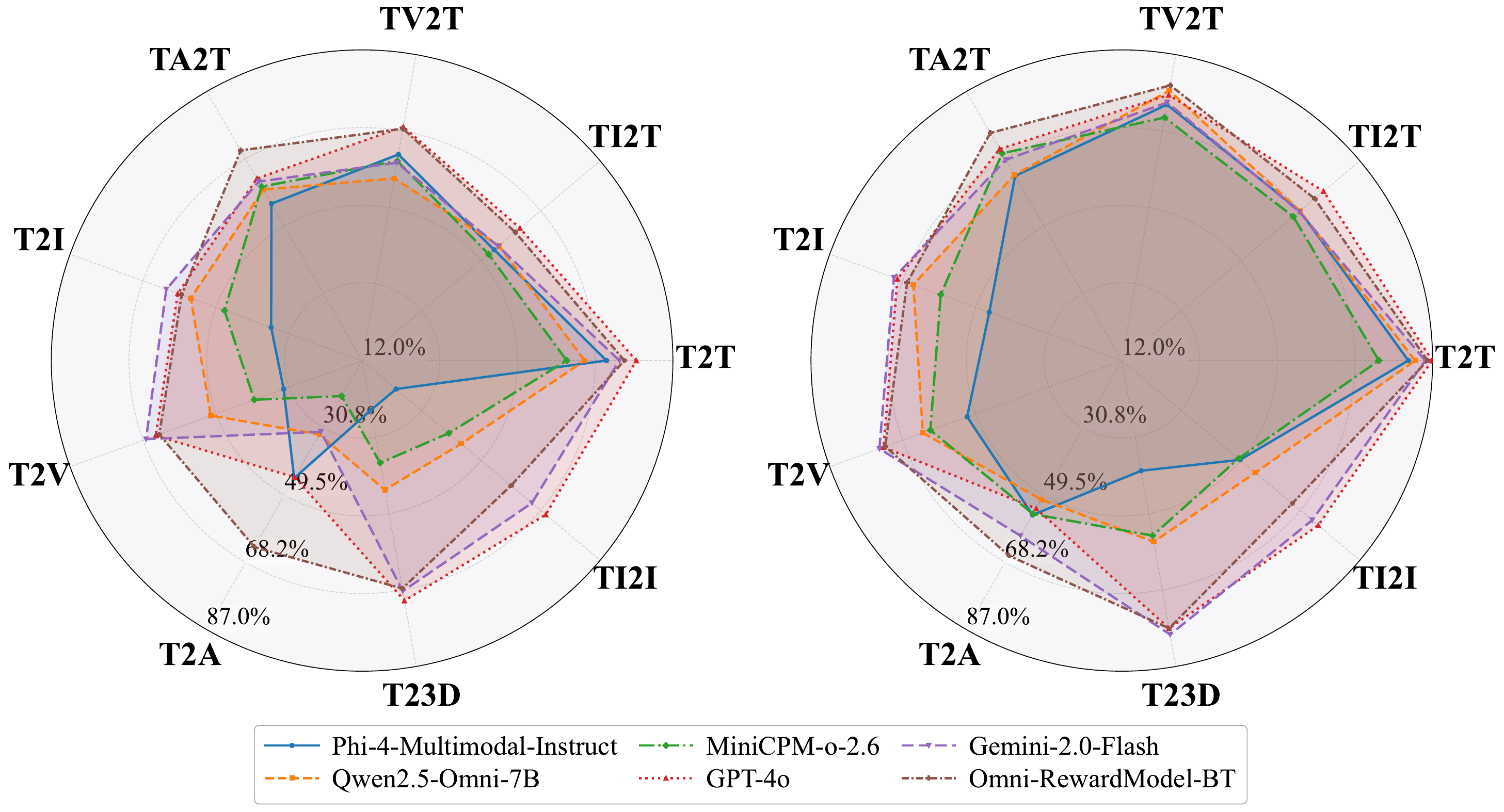}
    \caption{Performance of open-source models, closed-source models, and our proposed model on the nine tasks in \texttt{Omni-RewardBench}, with results under \textit{w/ Tie} (\textbf{left}) and \textit{w/o Tie} (\textbf{right}).
}
    \label{fig:radar_plot}
\end{figure}

\textbf{Specialized Reward Models.} We evaluate several custom RMs that are specifically trained on particular reward modeling tasks.
PickScore \citep{pick} and HPSv2 \citep{hps} are CLIP-based scoring functions trained for image generation tasks. InternLM-XComposer2.5-7B-Reward \citep{IXC} broadens the scope to multimodal understanding tasks that cover text, images, and videos. UnifiedReward \citep{wang2025unified} further incorporates both generation and understanding capabilities across image and video modalities. 
\vspace{-5pt}

\subsection{Implementation Details} 
\vspace{-5pt}

We conduct experiments under two evaluation settings: \textit{w/o Ties} and \textit{w/ Ties}.
For the \textit{w/o Ties} setting, we exclude all samples labeled as tie and require the model to choose the preferred response from $\{y_1, y_2\}$.
For the \textit{w/ Ties} setting, the model is required to select from $\{y_1, y_2, \text{tie}\}$.
Accuracy is used as the primary evaluation metric.
For generative RMs, we adopt a pairwise format where the model first generates explicit critiques for both responses, and then produces a final preference decision.
Prompt templates for generative RMs are detailed in Appendix~\ref{appendix:prompt_template}.
For discriminative RMs, we follow prior work \citep{DBLP:conf/emnlp/DeutschFF23} and define the \textit{w/ Ties} accuracy  as the maximum three-class classification accuracy obtained by varying the tie threshold.
More details are shown in Appendix \ref{Implementation Details}.

\vspace{-5pt}

\subsection{Evaluation Results on Omni-RewardBench}
\vspace{-5pt}

The evaluation results on \texttt{Omni-RewardBench} are shown in Table \ref{evaluation_result_w_tie}, Table \ref{evaluation_result_w_o_tie} and Figure \ref{fig:radar_plot}.

\textbf{Limited Performance of Current RMs.} The overall performance of current RMs remains limited, particularly under the \textit{w/ Ties} setting.
For instance, the strongest proprietary model, Claude 3.5 Sonnet, achieves an accuracy of \textbf{66.54\%}, while the best-performing open-source model, Gemma-3 27B, follows closely with \textbf{65.12\%}. 
In contrast, specialized reward models perform less competitively, with the most capable one, UnifiedReward1.5, achieving only \textbf{59.69\%} accuracy.
These results reveal that current RMs remain inadequate for omni-modal and free-form preference reward modeling, reinforcing the need for more capable and generalizable approaches.

\textbf{Modality Imbalance across Various Tasks.} As shown in Figure~\ref{fig:radar_plot}, task-level performance varies considerably, with up to a 28.37\% gap across modalities.
In particular, tasks like T2A, T23D, and TI2I perform notably worse, highlighting a persistent modality imbalance, as current reward models primarily focus on text and image, while modalities such as audio and 3D remain underexplored.

\textbf{Strong Performance of Omni-RewardModel.} 
\texttt{Omni-RewardModel-BT} achieves strong performance on the \texttt{Omni-RewardBench}, attaining \textbf{73.68\%} accuracy under the \textit{w/o Ties} setting and \textbf{65.36\%} accuracy under the \textit{w/ Ties} setting.
It also generalizes well to unseen modalities, achieving SOTA performance on TA2T and T2A tasks.
\texttt{Omni-RewardModel-R1} also surpasses existing specialized RMs in performance while providing better interpretability via explicit reasoning.

\vspace{-5pt}

\subsection{Evaluation Results on General Reward Modeling Benchmarks}

\begin{table}[t]
\centering
\caption{Evaluation results on VL-RewardBench.}
\label{VL-RewardBench}
 \resizebox{0.7\linewidth}{!}{
\begin{tabular}{lccccc}
\toprule
\textbf{Models}                    & \textbf{General} & \textbf{Hallucination} & \textbf{Reasoning} & \textbf{Overall Acc} & \textbf{Macro Acc} \\ \midrule
\multicolumn{6}{c}{\textit{Open-Source Models}}                                     \\ 
LLaVA-OneVision-7B-ov & 32.2          & 20.1 & 57.1          & 29.6 & 36.5 \\
Molmo-7B              & 31.1          & 31.8 & 56.2          & 37.5 & 39.7 \\
InternVL2-8B          & 35.6          & 41.1 & 59.0          & 44.5 & 45.2 \\
Llama-3.2-11B         & 33.3          & 38.4 & 56.6          & 42.9 & 42.8 \\
Pixtral-12B           & 35.6          & 25.9 & 59.9          & 35.8 & 40.4 \\
Molmo-72B             & 33.9          & 42.3 & 54.9          & 44.1 & 43.7 \\
Qwen2-VL-72B          & 38.1          & 32.8 & 58.0          & 39.5 & 43.0 \\
NVLM-D-72B            & 38.9          & 31.6 & 62.0          & 40.1 & 44.1 \\
Llama-3.2-90B         & 42.6          & 57.3 & 61.7          & 56.2 & 53.9 \\ \midrule
\multicolumn{6}{c}{\textit{Proprietary Models}}                                     \\ 
Gemini-1.5-Flash      & 47.8          & 59.6 & 58.4          & 57.6 & 55.3 \\
Gemini-1.5-Pro        & 50.8          & 72.5 & 64.2          & 67.2 & 62.5 \\
Claude-3.5-Sonnet     & 43.4          & 55.0 & 62.3          & 55.3 & 53.6 \\
GPT-4o-mini           & 41.7          & 34.5 & 58.2          & 41.5 & 44.8 \\
GPT-4o                & 49.1          & 67.6 & \textbf{70.5} & 65.8 & 62.4 \\ \midrule
\multicolumn{6}{c}{\textit{Specialized Models}}                                     \\ 
LLaVA-Critic-8B       & 54.6          & 38.3 & 59.1          & 41.2 & 44.0 \\
IXC-2.5-Reward        & \textbf{84.7} & 62.5 & 62.9          & 65.8 & 70.0 \\
UnifiedReward         & 60.6          & 78.4 & 60.5          & 66.1 & 66.5 \\
Skywork-VL-Reward         & 66.0          & 80.0	 & 61.0          & 73.1 & 69.0 \\
\rowcolor{gray!15}
\texttt{Omni-RewardModel-R1} & 71.9             & 90.2          & 59.0               & 69.6        & 73.7      \\ 
\rowcolor{gray!15}
\texttt{Omni-RewardModel-BT} & 81.5             & \textbf{94.2}          & 60.4               & \textbf{76.3}        & \textbf{78.7}      \\ \bottomrule
\end{tabular}}
\end{table}

\vspace{-5pt}
We further evaluate \texttt{Omni-RewardModel} on other widely-used RM benchmarks to assess its ability to model general human preferences.
VL-RewardBench \citep{vlrewardbench} evaluates multimodal RMs across general multimodal queries, visual hallucination detection, and complex reasoning tasks.
Multimodal RewardBench \citep{MultimodalRewardBench} covers six domains: general correctness, preference, knowledge, reasoning, safety, and visual question-answering.
In Table \ref{VL-RewardBench}, \texttt{Omni-RewardModel} achieves SOTA performance on VL-RewardBench, with an accuracy of \textbf{76.3\%}.
On Multimodal RewardBench (Table \ref{Multimodal-RewardBench}), \texttt{Omni-RewardModel} also matches the performance of Claude 3.5 Sonnet.

\begin{table}[t]
\centering
\caption{Evaluation results on Multimodal RewardBench.}
\label{Multimodal-RewardBench}
 \resizebox{\linewidth}{!}{
\begin{tabular}{lccccccccc}
\toprule
\multirow{2}{*}{\textbf{Model}} & \multirow{2}{*}{\textbf{Overall}} &
\multicolumn{2}{c}{\textbf{General}} &
\multirow{2}{*}{\textbf{Knowledge}} &
\multicolumn{2}{c}{\textbf{Reasoning}} &
\multicolumn{2}{c}{\textbf{Safety}} & \multirow{2}{*}{\textbf{VQA}} \\
& & \textbf{Correctness} & \textbf{Preference} & & \textbf{Math} & \textbf{Coding} & \textbf{Bias} & \textbf{Toxicity} & \\ \midrule
\multicolumn{10}{c}{\textit{Open-Source Models}}                                                                                                   \\
Llama-3.2-90B-Vision & 62.4          & 60.0          & 68.4          & 61.2 & 56.3          & 53.1 & 52.0          & 51.8 & 77.1          \\
Aria                          & 57.3          & 59.5          & 63.5          & 55.5 & 50.3          & 54.2 & 46.1          & 54.4 & 64.2          \\
Molmo-7B-D-0924               & 54.3          & 56.8          & 59.4          & 54.6 & 50.7          & 53.4 & 34.8          & 53.8 & 60.3          \\
Llama-3.2-11B-Vision & 52.4          & 57.8          & 65.8          & 55.5 & 50.6          & 51.7 & 20.9          & 50.4 & 55.8          \\
Llava-1.5-13B                 & 48.9          & 53.3          & 55.2          & 50.5 & 53.5          & 49.3 & 20.1          & 50.0 & 51.8          \\ \midrule
\multicolumn{10}{c}{\textit{Proprietary Models}}                                                                                                   \\
Claude 3.5 Sonnet &
  \textbf{72.0} &
  62.6 &
  67.8 &
  \textbf{73.9} &
  68.6 &
  \textbf{65.1} &
  76.8 &
  \textbf{60.6} &
  85.6 \\
Gemini 1.5 Pro                & \textbf{72.0} & 63.5          & 67.7          & 66.3 & 68.9          & 55.5 & \textbf{94.5} & 58.2 & \textbf{87.2} \\
GPT-4o                        & 71.5          & 62.6          & \textbf{69.0} & 72.0 & 67.6          & 62.1 & 74.8          & 58.8 & \textbf{87.2} \\ \midrule
\multicolumn{10}{c}{\textit{Specialized Models}}                                                                                                   \\
\rowcolor{gray!15}
\texttt{Omni-RewardModel-BT}           & 70.5          & \textbf{71.3} & 58.4          & 66.7 & \textbf{71.0} & 48.5 & 79.3          & -    & 85.1          \\ \bottomrule
\end{tabular}
}
\end{table}

\vspace{-10pt}

\section{Analysis}

\vspace{-5pt}

\begin{table}[t]
\centering
\caption{Ablation results on \texttt{Omni-RewardBench} under the \textit{w/ Tie} setting.}
\label{evaluation_ablation_1}
 \resizebox{\linewidth}{!}{

\begin{tabular} 
{
>{\columncolor[HTML]{FDFAF6}}l 
>{\columncolor[HTML]{fbf4f5}}c  
>{\columncolor[HTML]{fef0e7}}c  
>{\columncolor[HTML]{fffef1}}c  
>{\columncolor[HTML]{f3f7ec}}c  
>{\columncolor[HTML]{fff5f5}}c  
>{\columncolor[HTML]{fffaf3}}c  
>{\columncolor[HTML]{f3f7ec}}c  
>{\columncolor[HTML]{f5fcfe}}c  
>{\columncolor[HTML]{f9f8ff}}c  
>{\columncolor[HTML]{CDF5FD}}c }

\toprule

\textbf{Model} & \textbf{T2T} & \textbf{TI2T} & \textbf{TV2T} & \textbf{TA2T} & \textbf{T2I} & \textbf{T2V} & \textbf{T2A} & \textbf{T23D} & \textbf{TI2I} & \textbf{Overall} \\
\midrule

MiniCPM-o-2.6 & 61.39 & 51.89 & 60.95 & 60.50 & 47.35 & 39.70 & 21.90 & 37.09 & 39.30 & 46.67 \\

\ \ \ \ w/ T2T & 74.30 & 54.73 & 66.37 & 69.75 & 45.38 & 43.86 & 55.96 & 49.67 & 54.15 & 57.13 \\

\ \ \ \ w/ TI2T & 74.54 & 59.62 & 66.82 & 69.75 & 41.45 & 48.77 & 61.31 & 51.00 & 56.33 & 58.84 \\

\ \ \ \ w/ T2I \& T2V & 52.28 & 45.83 & 51.47 & 59.38 & \textbf{58.93} & \textbf{64.84} & 56.93 & 67.55 & \textbf{60.26} & 57.50 \\

\ \ \ \ w/ Full  & \textbf{75.30} & \textbf{60.23} & \textbf{68.85} & \textbf{70.59} & 58.35 & 64.08 & \textbf{63.99} & \textbf{67.88} & 58.95 & \textbf{65.36} \\

\ \ \ \ w/ Preference-Only & 54.92 & 49.80 & 64.79 & 55.74 & 59.14 & 61.06 & 64.00 & 64.90 & 53.71 & 58.67 \\

\bottomrule
\end{tabular}
}
\end{table}

In this section, we analyze the impact of training data composition in \texttt{Omni-RewardData} and examine the correlations among model performances across tasks in \texttt{Omni-RewardBench}. We further investigate the roles of CoT reasoning, free-form criteria, and scoring strategy in Appendix \ref{appendix_additional_analysis}.

\subsection{Impact of Training Data Composition}
\vspace{-5pt}

\begin{wrapfigure}{R}{0.4\linewidth} 
    \centering
    \vspace{-34pt}
    \includegraphics[width=1\linewidth]{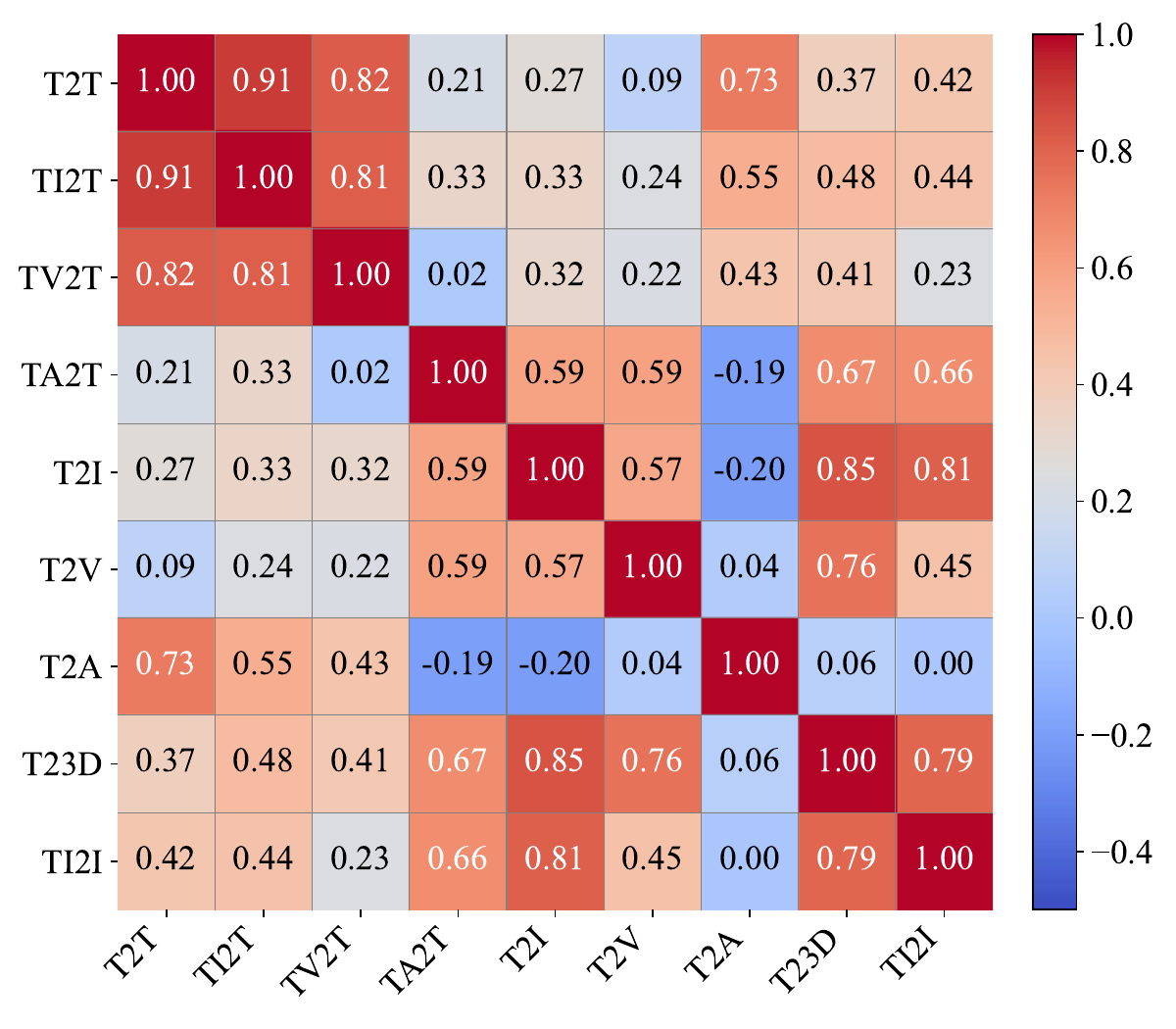}
    \vspace{-24pt}
    \caption{Performance correlation across various tasks in \texttt{Omni-RewardBench}.}
    \vspace{-20pt}
    \label{fig:corr_map}
\end{wrapfigure}

We examine the impact of training data composition on \texttt{Omni-RewardModel}, focusing on two key factors: the use of mixed multimodal data and the incorporation of instruction-tuning.
First, to assess the role of mixed multimodal data, we train MiniCPM-o-2.6 separately on (1) T2T, (2) TI2T, and (3) T2I and T2V data.
As shown in Tables~\ref{evaluation_ablation_1} and~\ref{evaluation_ablation_2}, while training on a single modality yields only marginal improvements, using mixed multimodal data leads to significantly better generalization across tasks.
Second, to assess the role of instruction-tuning data, we remove this type of data and train MiniCPM-o-2.6 using only the general preference data in \texttt{Omni-RewardData}. 
This leads to a clear drop in performance, highlighting the importance of instruction-tuning for RMs.

\vspace{-10pt}

\subsection{Correlation of Performance on Different Tasks}
\vspace{-5pt}

We analyze RM performance across nine tasks and reveal a significant degree of performance correlation among related tasks.
Specifically, we compute the Pearson correlation coefficients between tasks based on RM performance across the nine tasks in \texttt{Omni-RewardBench} and present the inter-task correlations as shown in Figure \ref{fig:corr_map}. 
We can observe that the performance correlations among understanding tasks, including text, image, and video understanding, are notably strong, with Pearson coefficients ranging from 0.8 to 0.9. 
Similarly, generation tasks such as video, 3D, and image generation also exhibit relatively high correlations, with scores mostly between 0.7 and 0.8.
These correlations suggest that RMs capture shared patterns within understanding and generation tasks, demonstrating generalization potential across modalities.

\vspace{-5pt}
\section{Related Work}
\vspace{-5pt}

\subsection{Multimodal Reward Model}
\vspace{-5pt}

Reinforcement learning from human feedback (RLHF) \citep{DBLP:journals/corr/abs-1909-08593, instructgpt, dpo, alignanything, rwku, yu2025aligning} has emerged as an effective approach for aligning MLLMs with human preferences, thereby enhancing multimodal understanding \citep{DBLP:journals/corr/abs-2404-01258, miadpo, OmniAlign-V}, reducing hallucinations \citep{DBLP:conf/acl/SunSCLLSGGWYKD24, DBLP:conf/cvpr/YuYZHHCHL0024, rlaifv, DBLP:conf/acl/JinCY0XLJ0024}, improving reasoning ability \citep{mpo, DBLP:journals/corr/abs-2503-06749}, and increasing safety \citep{mmrlhf, DBLP:conf/aaai/YuanJC0LZ25}.
Moreover, alignment is also beneficial for multimodal generation tasks, such as text-to-image generation \citep{DBLP:journals/corr/abs-2302-12192, DBLP:conf/cvpr/LiangHLLKCSPYYK24, ImageReward} and text-to-video generation \citep{DBLP:journals/corr/abs-2412-02617, lift, VideoReward, DBLP:journals/corr/abs-2502-10248}, by improving generation quality and controllability.
In the alignment process, reward models are crucial for modeling human preferences and providing feedback signals that guide the model toward generating more desirable and aligned outputs.
However, most existing reward models \citep{DBLP:journals/corr/abs-2110-14168, DBLP:conf/acl/WangLSXDLCWS24, skywork} primarily focus on text-to-text generation tasks, offering limited support for multimodal inputs and outputs.
Recently, an increasing number of reward models have been proposed to support multimodal tasks.
For example, PickScore \citep{DBLP:conf/cvpr/LiangHLLKCSPYYK24}, ImageReward \citep{ImageReward}, and HPS \citep{hps, hpdv2} are designed to evaluate the quality of text-to-image generation.
VisionReward \citep{VisionReward}, VideoReward \citep{VideoReward}, and VideoScore \citep{VideoScore} focus on assessing text-to-video generation.
LLaVA-Critic \citep{llavacritic} and IXC-2.5-Reward \citep{IXC} aim to align vision-language models by evaluating their instruction following and reasoning capabilities.
UnifiedReward \citep{wang2025unified} is the first unified reward model for assessing both visual understanding and generation tasks.
However, existing multimodal reward models remain inadequate for fully omni-modal scenarios, 

\vspace{-5pt}
\subsection{Reward Model Evaluation}
\vspace{-5pt}
As the diversity of reward models expands, a growing number of benchmarks are emerging to address the need for evaluation \citep{ RAG-RewardBench, ProcessBench, DBLP:journals/corr/abs-2503-07478, Agent-RewardBench}.
RewardBench \citep{rewardbench} is the first comprehensive framework for assessing RMs in chat, reasoning, and safety domains.
Furthermore, RMB \citep{RMB} broadens the evaluation scope by including 49 real-world scenarios.
RM-Bench \citep{rmbench} is designed to evaluate RMs based on their sensitivity to subtle content differences and style biases.
In the multimodal domain, several benchmarks have been proposed to evaluate reward models for image generation, such as MJ-Bench \citep{mjbench} and GenAI-Bench \citep{jiang2024genai}.
For video generation, VideoGen-RewardBench \citep{VideoReward} provides a suitable benchmark for assessing visual quality, motion quality, and text alignment.
More broadly, VL-RewardBench \citep{vlrewardbench} and Multimodal RewardBench \citep{MultimodalRewardBench} have been proposed to evaluate reward models for vision-language models. 
Extending further, AlignAnything \citep{alignanything} collects large-scale human preference data across modalities for post-training alignment and evaluates the general capabilities of omni-modal models.
Meanwhile, in text-to-text generation tasks, several recent studies such as PRP \citep{DBLP:conf/nips/PitisXRS24}, HelpSteer2-Preference \citep{wang2024helpsteer2}, and GRM \citep{deepseek_grm} have started to focus on fine-grained reward modeling.
However, existing benchmarks lack a unified framework for evaluating reward models with respect to specific textual criteria across diverse multimodal scenarios.

\vspace{-5pt}

\section{Conclusion}

\vspace{-5pt}

In this paper, we present \texttt{Omni-Reward}, a unified framework for omni-modal reward modeling with free-form user preferences. To address the challenges of modality imbalance and preference rigidity in current RMs, we introduce three key components: (1) \texttt{Omni-RewardBench}, a  comprehensive RM benchmark spanning five modalities and nine diverse tasks; (2) \texttt{Omni-RewardData}, a large-scale multimodal preference dataset incorporating both general and instruction-tuning data; and (3) \texttt{Omni-RewardModel}, a family of discriminative and generative RMs with strong performance.

\section*{Ethics Statement}

This research involves human annotations to construct preference data. All annotation tasks were conducted by the authors of this paper, who participated voluntarily and with full knowledge of the study’s purpose, procedures, and intended use of the data. No external crowdsourcing or paid annotation platforms were employed. To safeguard research integrity and mitigate potential biases, detailed annotation protocols and quality control measures are documented in the Appendix \ref{echics_and_quality_control}.

The study does not involve sensitive personal data, human subjects outside of the annotation task, or applications that raise privacy, security, or legal concerns. 
We also follow the standard research ethics protocols of our institution, with explicit approval from the IRB, for all internal annotation efforts.
The research complies with the ICLR Code of Ethics, and no conflicts of interest or sponsorship concerns are associated with this work.

\section*{Reproducibility statement}

We have taken extensive measures to ensure the reproducibility of our results. All implementation details of the proposed \texttt{Omni-Reward} framework, including architectures, training procedures, and evaluation protocols, are described in the main paper and further elaborated in the Appendix. To support future research, we will release \texttt{Omni-RewardBench}, \texttt{Omni-RewardData}, and \texttt{Omni-RewardModel} as part of a comprehensive open-source package. 
All assets we provide are licensed under the Creative Commons Attribution Non Commercial 4.0 International License (CC BY-NC 4.0).
In addition, complete data processing steps and annotation protocols are documented in the Appendix. These efforts are intended to enable the community to replicate our experiments and build upon our findings.

\newpage
\bibliographystyle{iclr2026_conference}
\bibliography{iclr2026_conference}

\newpage


\appendix

\section{LLM Usage Statement}

LLMs were used solely as auxiliary tools for grammar checking and language polishing. They did not contribute to the generation of research ideas, the design of experiments, the development of methodologies, data analysis, or any substantive aspects of the research. All scientific content, conceptual contributions, and experimental results are entirely the work of the authors. The authors take full responsibility for the contents of this paper.

\section{Limitations}
\label{Limitations}
In this section, we outline some limitations of our work. (1) Our \texttt{Omni-RewardBench} is a benchmark consisting of several thousand human-labeled preference pairs. Its current scale may not be sufficient to support evaluations at much larger magnitudes, such as those involving millions of examples. (2) While our benchmark covers nine distinct task types across different modalities, current task definitions remain relatively coarse, and further fine-grained categorization within each task type is desired. (3) The current preference data is limited to single-turn interactions and does not capture multi-turn conversational preferences, which are increasingly important for modeling real-world dialogue scenarios. (4) The reinforcement learning technique in training the \texttt{Omni-RewardModel-R1} is limited to a preliminary exploration, and further investigation is needed.
(5) Incorporating additional modalities such as thermal, radar, tabular data, and time-series data would further enhance the scope and utility of our benchmark.

\vspace{-5pt}

\section{Broader Impacts}
\label{Broader Impacts}

\vspace{-5pt}

Some preference pairs in \texttt{Omni-Reward} may contain offensive, inappropriate, or otherwise sensitive prompts and responses, as they are intended to reflect real-world scenarios.
We recommend that users exercise caution and apply their own ethical guidelines when using the dataset.

\clearpage

\vspace{-5pt}

\section{Annotation Details}
\label{Annotation Details}

\vspace{-5pt}

\subsection{Construction Workflow}

\begin{figure*}[!h]
\centering
\resizebox{\linewidth}{!}{
\includegraphics[]{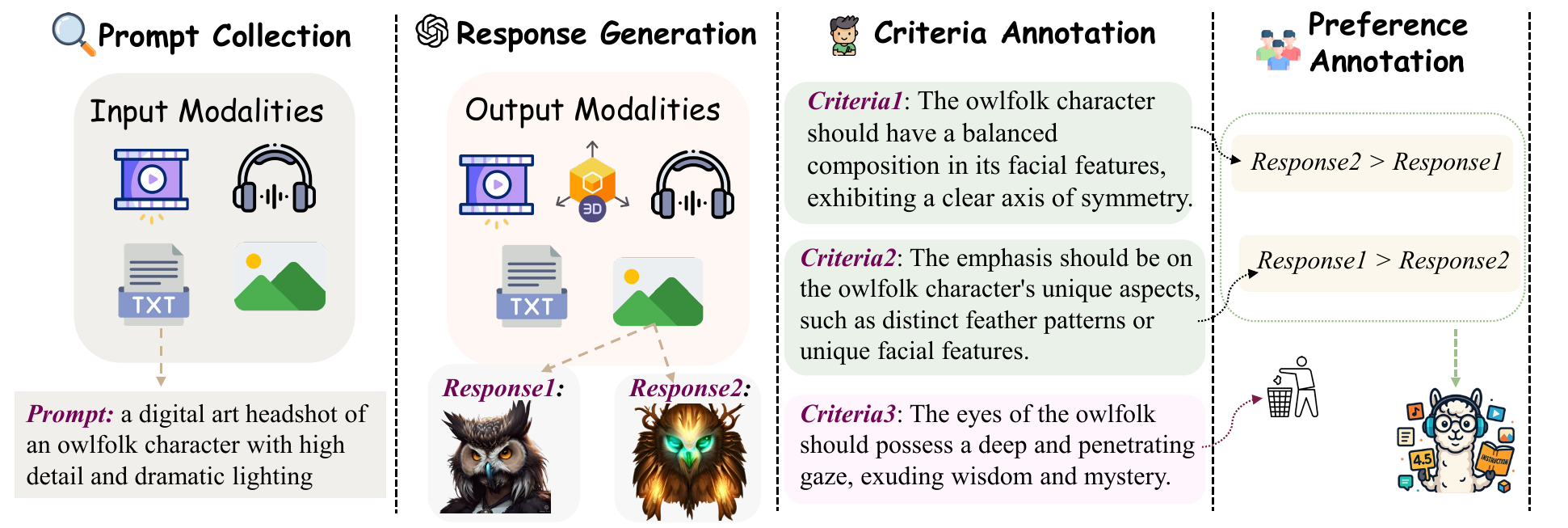} 
}
\caption{Construction workflow of \texttt{Omni-RewardBench}.}
\label{dataset_construction_workflow}
\end{figure*}

\vspace{-5pt}

\subsection{Annotation Guideline}

\begin{tcolorbox}[size=title,opacityfill=0.05,breakable]

\textbf{1. Objective}

This annotation task aims to identify and label evaluation dimensions under which one model response (Response A) is preferred over another (Response B), given a specific task instance (e.g., text-to-image generation, video understanding, or text-to-audio generation). The annotated dataset will serve as a foundation for building robust evaluation benchmarks that reflect nuanced human preferences across different modalities and task types.

\textbf{2. Task Definition} 

Each data instance consists of the following components:

A task description (e.g., a prompt or instruction corresponding to a specific task category such as image generation or video analysis),

Two model responses, denoted as Response A and Response B.

Annotators are expected to analyze the responses and determine which aspects make one response superior to the other, focusing on concrete and interpretable evaluation dimensions (e.g., relevance, coherence, visual quality).

\textbf{3. Annotation Procedure}

The annotation process involves the following steps:

(1) Carefully read the task description and understand the intended objective.

(2) Examine Response A and Response B in the context of the given task.

(3) Write one or more evaluation dimension descriptions using fluent, complete English sentences. Each sentence should define a specific, human-interpretable dimension along which the two responses can be meaningfully compared.

(4) For each evaluation dimension that you articulate, assign a comparative label among the following three:

Response A is better,

Response B is better,

Both responses are equivalent.

\end{tcolorbox}

\clearpage

\subsection{Annotation Platform}

\begin{figure*}[!h]
    \centering
    \resizebox{0.95\linewidth}{!}{
\includegraphics[]{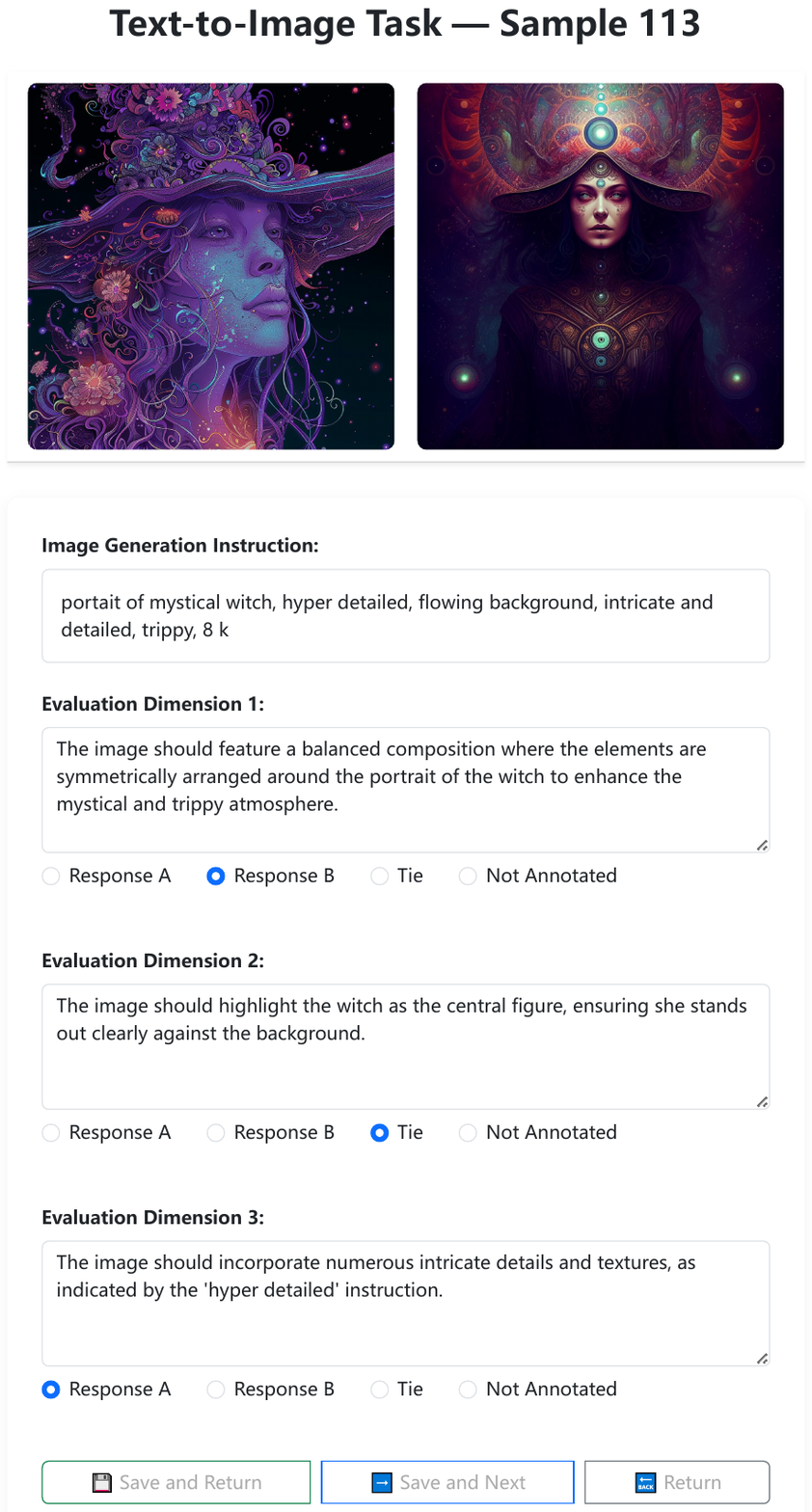} 
}
    \caption{Annotation platform for human annotators.}
    \label{fig:annotation_platform}
\end{figure*}

\clearpage

\section{Ethics and Quality Control}
\label{echics_and_quality_control}

\subsection{Ethics}
We confirm that all annotations were conducted voluntarily by the authors of this paper, who were fully informed about the nature and purpose of the task, their rights, and how the data would be used. We also follow the standard research ethics protocols of our institution, with explicit approval from the IRB, for all internal annotation efforts.

\subsection{Quality Control}
As illustrated in Figure \ref{dataset_construction_workflow}, our annotation pipeline consists of two key stages: Criteria Annotation and Preference Annotation. Throughout these two stages, we removed a total of $38\%$ of the samples to ensure data quality.

\begin{itemize}
    \item \textbf{Criteria Annotation.} We filtered out $23\%$ of the samples whose criteria were deemed either too vague or overly specific, as part of our quality control on preference criteria. Such criteria would undermine the overall consistency and utility of the preference data.
    
    \item \textbf{Preference Annotation.} We further removed $15\%$ of the samples due to disagreements among annotators, where no consensus could be reached on the preferred output. To quantify inter-rater reliability, we report Krippendorff’s alpha of 0.701, indicating substantial agreement among annotators.
\end{itemize}

The annotation was carried out by a small group of PhD students. Despite the resource-intensive nature of the task, we undertook extensive measures, as documented in Appendix \ref{Annotation Details}, to safeguard annotation consistency and mitigate potential biases. These procedures collectively ensured that the final dataset is both ethically collected and of high quality.

Moreover, unlike broad and subjective preferences such as helpfulness or harmlessness, our benchmark provides explicit and well-defined textual criteria for each annotation instance. This design choice reduces the risk of ambiguity and limits the impact of cultural or individual variation in interpretation, thereby minimizing the potential issues arising from a lack of demographic diversity among annotators.

\newpage

\section{Dataset Statistics}

\subsection{Benchmark Comparison}

Table \ref{dataset_comparison} presents a detailed comparison between \texttt{Omni-RewardBench} and existing reward modeling benchmarks. 
While prior benchmarks often focus on a narrow range of modalities or task types, \texttt{Omni-RewardBench} provides the most comprehensive coverage, spanning nine tasks across five modalities: text, image, video, audio, and 3D.
Moreover, \texttt{Omni-RewardBench} uniquely supports free-form preference annotations, allowing more expressive and fine-grained evaluation criteria compared to the binary preferences used in most existing datasets.
Notably, Table \ref{dataset_comparison} shows that AlignAnything bears similarity to \texttt{Omni-RewardBench}. As an influential contribution, it has inspired several aspects of \texttt{Omni-Reward}, particularly the notion of any-to-any alignment. Nevertheless, a key distinction exists: AlignAnything concentrates on aligning omni-modal models to enhance their capabilities across diverse input–output modalities, introducing EvalAnything to assess the performance of the aligned models. By contrast, our work emphasizes reward modeling within the alignment pipeline, with \texttt{Omni-RewardBench} designed to directly evaluate reward models by testing whether their inferred preferences align with human judgments under specified textual criteria.

We compare the performance of ten models on OmniRewardBench and VLRewardBench, obtaining a Spearman correlation coefficient of 0.4572 between their rankings. This indicates that incorporating additional modalities and free-form criteria differentiates our benchmark from previous ones.

\begin{table}[!h]
\centering
\caption{The comparison between \texttt{Omni-RewardBench} and other reward modeling benchmarks.}
 \resizebox{\linewidth}{!}{
\begin{tabular}{ccccccccccccc}
\toprule
                                     &                                   & \multicolumn{9}{c}{\textbf{Tasks}}                                                                                                                                                                                                               &                                                 &                                       \\
\multirow{-2}{*}{\textbf{Benchmark}} & \multirow{-2}{*}{\textbf{\#Size}} & \textbf{T2T}             & \textbf{TI2T}            & \textbf{TV2T}            & \textbf{TA2T}            & \textbf{T2I}             & \textbf{T2V}             & \textbf{T2A}             & \textbf{T23D}            & \textbf{TI2I}            & \multirow{-2}{*}{\textbf{\begin{tabular}[c]{@{}c@{}}Free-Form \\ Preference\end{tabular}}} & \multirow{-2}{*}{\textbf{Annotation}} \\ \midrule

RewardBench \citep{rewardbench}                          & 2,985                             & {\color[HTML]{00B050} \checkmark} & {\color[HTML]{FF0000} $\times$} & {\color[HTML]{FF0000} $\times$} & {\color[HTML]{FF0000} $\times$} & {\color[HTML]{FF0000} $\times$} & {\color[HTML]{FF0000} $\times$} & {\color[HTML]{FF0000} $\times$} & {\color[HTML]{FF0000} $\times$} & {\color[HTML]{FF0000} $\times$} & {\color[HTML]{FF0000} $\times$}                        & Human                                 \\
RPR \citep{DBLP:conf/nips/PitisXRS24}                          & 10,167                             & {\color[HTML]{00B050} \checkmark} & {\color[HTML]{FF0000} $\times$} & {\color[HTML]{FF0000} $\times$} & {\color[HTML]{FF0000} $\times$} & {\color[HTML]{FF0000} $\times$} & {\color[HTML]{FF0000} $\times$} & {\color[HTML]{FF0000} $\times$} & {\color[HTML]{FF0000} $\times$} & {\color[HTML]{FF0000} $\times$} & {\color[HTML]{00B050} \checkmark}                        & GPT                                 \\
RM-Bench \citep{rmbench}                         & 1,327                             & {\color[HTML]{00B050} \checkmark} & {\color[HTML]{FF0000} $\times$} & {\color[HTML]{FF0000} $\times$} & {\color[HTML]{FF0000} $\times$} & {\color[HTML]{FF0000} $\times$} & {\color[HTML]{FF0000} $\times$} & {\color[HTML]{FF0000} $\times$} & {\color[HTML]{FF0000} $\times$} & {\color[HTML]{FF0000} $\times$} & {\color[HTML]{FF0000} $\times$}                        & GPT                                 \\
MJ-Bench \citep{mjbench}                         & 4,069                             & {\color[HTML]{FF0000} $\times$} & {\color[HTML]{FF0000} $\times$} & {\color[HTML]{FF0000} $\times$} & {\color[HTML]{FF0000} $\times$} & {\color[HTML]{00B050} \checkmark} & {\color[HTML]{FF0000} $\times$}  & {\color[HTML]{FF0000} $\times$} & {\color[HTML]{FF0000} $\times$} & {\color[HTML]{FF0000} $\times$}  & {\color[HTML]{FF0000} $\times$}                        & Human                                 \\
GenAI-Bench \citep{jiang2024genai}                         & 9,810                             & {\color[HTML]{FF0000} $\times$} & {\color[HTML]{FF0000} $\times$} & {\color[HTML]{FF0000} $\times$} & {\color[HTML]{FF0000} $\times$} & {\color[HTML]{00B050} \checkmark} & {\color[HTML]{00B050} \checkmark} & {\color[HTML]{FF0000} $\times$} & {\color[HTML]{FF0000} $\times$} & {\color[HTML]{00B050} \checkmark} & {\color[HTML]{FF0000} $\times$}                        & Human                                 \\
VisionReward \citep{VisionReward}                        & 2,000                            & {\color[HTML]{FF0000} $\times$} & {\color[HTML]{FF0000} $\times$} & {\color[HTML]{FF0000} $\times$} & {\color[HTML]{FF0000} $\times$} & {\color[HTML]{00B050} \checkmark} & {\color[HTML]{00B050} \checkmark} & {\color[HTML]{FF0000} $\times$} & {\color[HTML]{FF0000} $\times$} & {\color[HTML]{FF0000} $\times$} & {\color[HTML]{FF0000} $\times$}                        & Human                                 \\
VideoGen-RewardBench \citep{VideoReward}                        & 26,457                            & {\color[HTML]{FF0000} $\times$} & {\color[HTML]{FF0000} $\times$} & {\color[HTML]{FF0000} $\times$} & {\color[HTML]{FF0000} $\times$} & {\color[HTML]{FF0000} $\times$} & {\color[HTML]{00B050} \checkmark} & {\color[HTML]{FF0000} $\times$} & {\color[HTML]{FF0000} $\times$} & {\color[HTML]{FF0000} $\times$} & {\color[HTML]{FF0000} $\times$}                        & Human                                 \\
MLLM-as-a-Judge \citep{MLLM-as-a-Judge}                      & 15,450                            & {\color[HTML]{FF0000} $\times$} & {\color[HTML]{00B050} \checkmark} & {\color[HTML]{FF0000} $\times$} & {\color[HTML]{FF0000} $\times$} & {\color[HTML]{FF0000} $\times$} & {\color[HTML]{FF0000} $\times$} & {\color[HTML]{FF0000} $\times$} & {\color[HTML]{FF0000} $\times$} & {\color[HTML]{FF0000} $\times$} & {\color[HTML]{FF0000} $\times$}                        & Human                                 \\
VL-RewardBench \citep{vlrewardbench}                       & 1,250                             & {\color[HTML]{FF0000} $\times$} & {\color[HTML]{00B050} \checkmark} & {\color[HTML]{FF0000} $\times$} & {\color[HTML]{FF0000} $\times$} & {\color[HTML]{FF0000} $\times$} & {\color[HTML]{FF0000} $\times$} & {\color[HTML]{FF0000} $\times$} & {\color[HTML]{FF0000} $\times$} & {\color[HTML]{FF0000} $\times$} & {\color[HTML]{FF0000} $\times$}                        & GPT+Human                             \\
Multimodal RewardBench \citep{MultimodalRewardBench}               & 5,211                             & {\color[HTML]{FF0000} $\times$} & {\color[HTML]{00B050} \checkmark} & {\color[HTML]{FF0000} $\times$} & {\color[HTML]{FF0000} $\times$} & {\color[HTML]{FF0000} $\times$} & {\color[HTML]{FF0000} $\times$} & {\color[HTML]{FF0000} $\times$} & {\color[HTML]{FF0000} $\times$} & {\color[HTML]{FF0000} $\times$} & {\color[HTML]{FF0000} $\times$}                        & Human                                 \\
MM-RLHF-RewardBench \citep{mmrlhf}               & 170                             & {\color[HTML]{FF0000} $\times$} & {\color[HTML]{00B050} \checkmark} & {\color[HTML]{00B050} \checkmark} & {\color[HTML]{FF0000} $\times$} & {\color[HTML]{FF0000} $\times$} & {\color[HTML]{FF0000} $\times$} & {\color[HTML]{FF0000} $\times$} & {\color[HTML]{FF0000} $\times$} & {\color[HTML]{FF0000} $\times$} & {\color[HTML]{FF0000} $\times$}                        & Human                                 \\
AlignAnything \citep{alignanything}                        & 20,000                            & {\color[HTML]{00B050} \checkmark} & {\color[HTML]{00B050} \checkmark} & {\color[HTML]{00B050} \checkmark} & {\color[HTML]{00B050} \checkmark} & {\color[HTML]{00B050} \checkmark} & {\color[HTML]{00B050} \checkmark} & {\color[HTML]{00B050} \checkmark} & {\color[HTML]{00B050} \checkmark} & {\color[HTML]{FF0000} $\times$} & {\color[HTML]{FF0000} $\times$}                        & GPT+Human                             \\
\texttt{Omni-RewardBench} (Ours)               & 3,725                             & {\color[HTML]{00B050} \checkmark} & {\color[HTML]{00B050} \checkmark} & {\color[HTML]{00B050} \checkmark} & {\color[HTML]{00B050} \checkmark} & {\color[HTML]{00B050} \checkmark} & {\color[HTML]{00B050} \checkmark} & {\color[HTML]{00B050} \checkmark} & {\color[HTML]{00B050} \checkmark} & {\color[HTML]{00B050} \checkmark} & {\color[HTML]{00B050} \checkmark}                        & Human          \\ \bottomrule                       
\end{tabular}
}
\label{dataset_comparison}
\end{table}

\subsection{Omni-RewardBench Statistics}

Due to the inherent difficulty of collecting high-quality data across multiple modalities, some imbalance in the distribution of preference pairs is unavoidable. While some imbalance remains, our dataset maintains a relatively balanced distribution across modalities, especially when compared to the significant disparities commonly observed in real-world data availability between modalities such as images and audio.

\begin{table}[!h]
\centering
\caption{Data statistics of \texttt{Omni-RewardBench}. The\textbf{ Avg. \#Tokens (Prompt)}, \textbf{Avg. \#Tokens (Response)}, and \textbf{Avg. \#Tokens (Criteria)} columns report the average number of tokens in the prompt, model-generated response, and human-written evaluation criteria, respectively, all measured using the tokenizer of Qwen2.5-VL-7B-Instruct. The \textbf{Prompt Source} column specifies where the prompts were collected from, while the \textbf{Model} column identifies which models were used to produce the corresponding responses.  The letters \textbf{“V”}, \textbf{“I”}, \textbf{“A”}, and \textbf{“D”} in the table stand for \textit{Video}, \textit{Image}, \textit{Audio}, and \textit{3D content}, respectively.}
 \resizebox{\linewidth}{!}{
\begin{tabular}{ccccccc}
\toprule
\textbf{Task} & \textbf{\#Pairs} & \textbf{\begin{tabular}[c]{@{}c@{}}Avg. \#Tokens\\ (Prompt)\end{tabular}} & \textbf{\begin{tabular}[c]{@{}c@{}}Avg. \#Tokens\\ (Response)\end{tabular}} & \textbf{\begin{tabular}[c]{@{}c@{}}Avg. \#Tokens\\ (Criteria)\end{tabular}} & \textbf{Prompt Source} & \textbf{\#Models} \\
\midrule
T2T   & 417 & 83.3 & 222.1 & 17.24 & RMB, RPR & 15 $^{\rm a}$ \\
TI2T  & 528 & 22.47 \& I & 104.66 & 15.71 & MIA-Bench, VLFeedback & 19 $^{\rm b}$ \\
TV2T  & 443 & 14.53 \& V & 133.42 & 14.69 & VCGBench-Diverse & 4 $^{\rm c}$ \\
TA2T  & 357 & 14.46 \& A & 77.83 & 21.85 & LTU & 2 $^{\rm d}$ \\
T2I   & 509 & 17.77 & I & 21.72 & HPDv2, Rapidata & 27 $^{\rm e}$ \\
T2V   & 529 & 9.61 & V & 23.29 & GenAI-Bench & 8$^{\rm f}$ \\
T2A   & 411 & 11.46 & A & 11.47 & Audio-alpaca & 1$^{\rm g}$ \\
T23D  & 302 & 14.32 & D & 30.21 & 3DRewardDB & 1$^{\rm h}$ \\
TI2I  & 229 & 7.89 \& I & I & 29.81 & GenAI-Bench & 10 $^{\rm i}$ \\
\midrule
Total & 3,725 & 27.29 & 134.50 & 20.67 & - & - \\
        \bottomrule
\end{tabular}
}

\vspace{0.5em}
\begin{flushleft}
\scriptsize{
$^{\rm a}$ Claude-3-5-Sonnet-20240620, Mixtral-8x7B-Instruct-v0.1, Vicuna-7B-v1.5, GPT-4o-mini-2024-07-18, Llama-2-7b-chat-hf, Mistral-7B-Instruct-v0.1, Claude-2.1, Gemini-1.5-Pro-Exp-0801, Llama-2-70b-chat-hf, Gemini-Pro, Qwen2-7B-Instruct, Claude-3-Opus-20240229, GPT-4 Turbo, Qwen1.5-1.8B-Chat, Claude-Instant-1.2. \\
$^{\rm b}$ GPT-4o, Gemini-1.5-Pro, Qwen2-VL-7B-Instruct, Claude-3-5-Sonnet-20240620, GPT-4o-mini, Qwen-VL-Chat, Llava1.5-7b, Gpt-4v, VisualGLM-6b, LLaVA-RLHF-13b-v1.5-336, MMICL-Vicuna-13B, LLaVA-RLHF-7b-v1.5-224, Instructblip-vicuna-7b, Fuyu-8b, Instructblip-vicuna-13b, Idefics-9b-instruct, Qwen-VL-Max-0809, Qwen-VL-plus, GLM-4v. \\
$^{\rm c}$ Qwen-VL-Max-0809, Qwen2-VL-7B-Instruct, Claude-3-5-Sonnet-20241022, GPT-4o. \\
$^{\rm d}$ Qwen-Audio, Gemini-2.0-Flash. \\
$^{\rm e}$ sdv2, VQGAN, SDXL-base-0.9, Cog2, CM, DALLE-mini, DALLE, DF-IF, ED, RV, flux-1.1-pro, Laf, LDM, imagen-3, DL, glide, OJ, MM, Deliberate, VD, sdv1, FD, midjourney-5.2, flux-1-pro, VQD, dalle-3, stable-diffusion-3. \\
$^{\rm f}$ LaVie, VideoCrafter2, ModelScope, AnimateDiffTurbo, AnimateDiff, OpenSora, T2VTurbo, StableVideoDiffusion. \\
$^{\rm g}$ Tango. \\
$^{\rm h}$ MVDream-SD2.1-Diffusers. \\
$^{\rm i}$ MagicBrush, SDEdit, InstructPix2Pix, CosXLEdit, InfEdit, Prompt2Prompt, Pix2PixZero, PNP, CycleDiffusion, DALL-E 2.
}
\end{flushleft}

    \label{appendix:dataset_statistics}
\end{table}

\begin{table}[]
    \centering
        \caption{Statistics of free-form criteria per preference pair in \texttt{Omni-RewardBench}.}
    \label{tab:number_of_criteria_per_pair}
    \begin{tabular}{lcccc}
    \toprule 
\textbf{Task} & \textbf{Mean} & \textbf{Median} & \textbf{Min} & \textbf{Max} \\
\midrule
T2T   & 2.7 & 2.0 & 1 & 6 \\

TI2T  & 2.8 & 3.0 & 1 & 6 \\
TV2T  & 2.6 & 3.0 & 1 & 6 \\
TA2T  & 2.8 & 3.0 & 1 & 3 \\

T2I   & 7.6 & 8.0 & 1 & 10 \\
T2V   & 4.4 & 5.0 & 1 & 5 \\
T2A   & 3.0 & 3.0 & 2 & 3 \\

T23D  & 4.2 & 4.0 & 1 & 6 \\
TI2I  & 2.0 & 2.0 & 1 & 4 \\
        \bottomrule

    \end{tabular}

\end{table}

\subsection{Omni-RewardData Statistics}

To mitigate potential systematic biases introduced by relying solely on GPT-4o, we incorporated a multi-model verification process \citep{DBLP:journals/corr/abs-2506-04141} to mitigate potential errors and biases introduced by GPT-4o during instruction generation. Notably, this filtering process is framed as a classification task, which is generally less complex and more robust than open-ended instruction generation, helping catch mistakes made by GPT-4o.

 \clearpage

\section{Implementation Details}
\label{Implementation Details}

For training \texttt{Omni-RewardModel-BT}, we use the LLaMA-Factory framework \footnote{\url{https://github.com/hiyouga/LLaMA-Factory}}. 
We adopt MiniCPM-o-2.6 as the base model and freeze the parameters of the vision encoder and audio encoder.
The model is trained for 2 epochs with a learning rate of 2e-6, weight decay of 1e-3, a cosine learning rate scheduler, and a warmup ratio of 1e-3.
For training \texttt{Omni-RewardModel-R1}, we use the EasyR1 framework \footnote{\url{https://github.com/hiyouga/EasyR1}}.
We adopt Qwen2.5-VL-7B-Instruct as the base model and freeze the parameters of the vision encoder.
The model is trained for 2 epochs with a learning rate of 1e-6, weight decay of 1e-2, and a rollout number of 6.
We use vllm \footnote{\url{https://github.com/vllm-project/vllm}} for open-source MLLM inference.
All experiments are conducted on 4×A100 80GB GPUs.
For evaluation, we compute the overall score by averaging the performance across all modalities supported by a given model.

\newpage
\section{Additional Experimental Results}
\label{appendix_additional_results}

\begin{figure}[h]
    \centering
    \includegraphics[width=\linewidth]{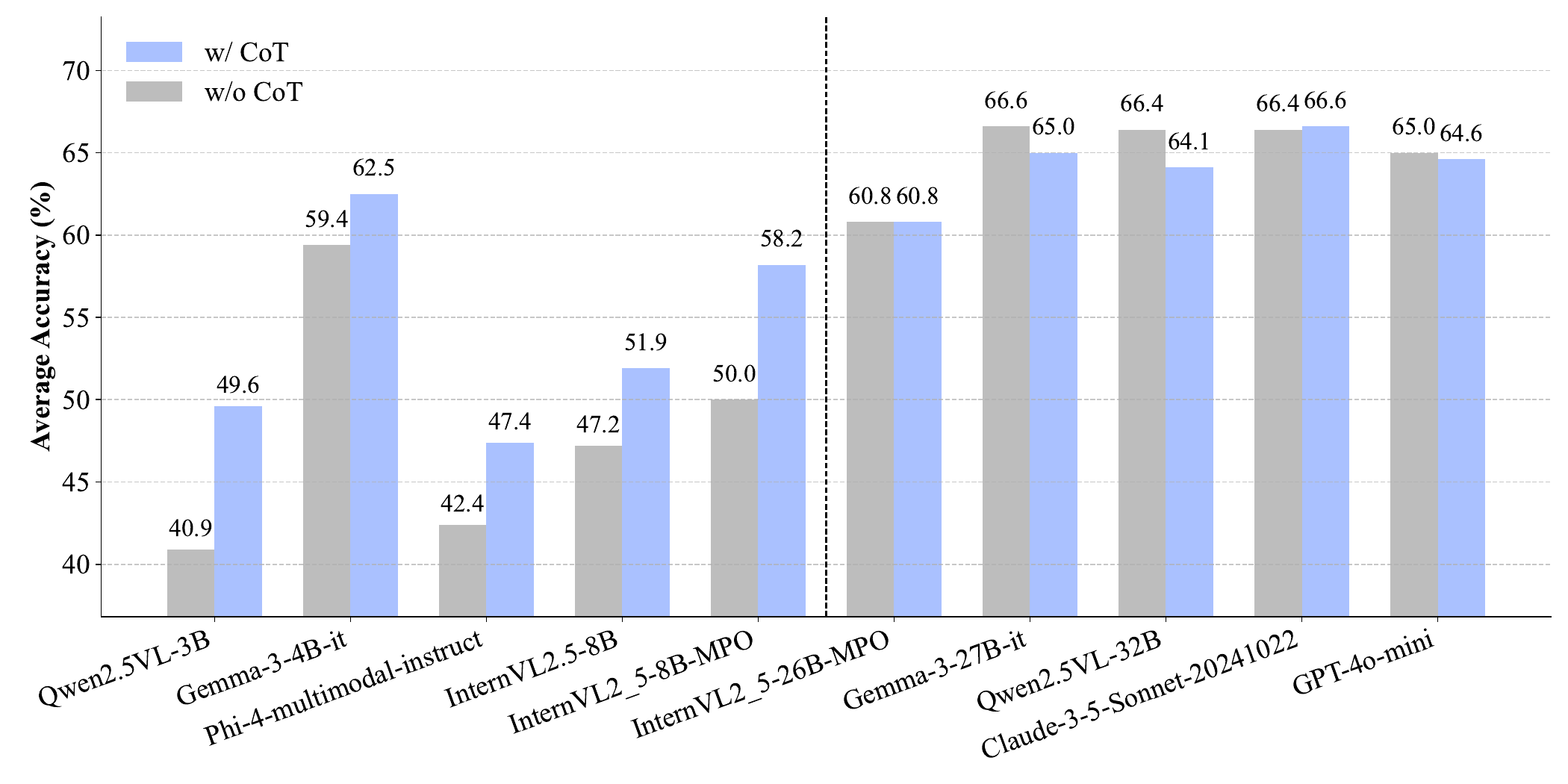}
                        
    \caption{Effect of CoT reasoning on \texttt{Omni-RewardBench} under \textit{w/ Tie} setting. }
                      
    \label{fig: cot_exp_visualize_with_tie}
\end{figure}

\begin{figure}[h]
    \centering
    \includegraphics[width=\linewidth]{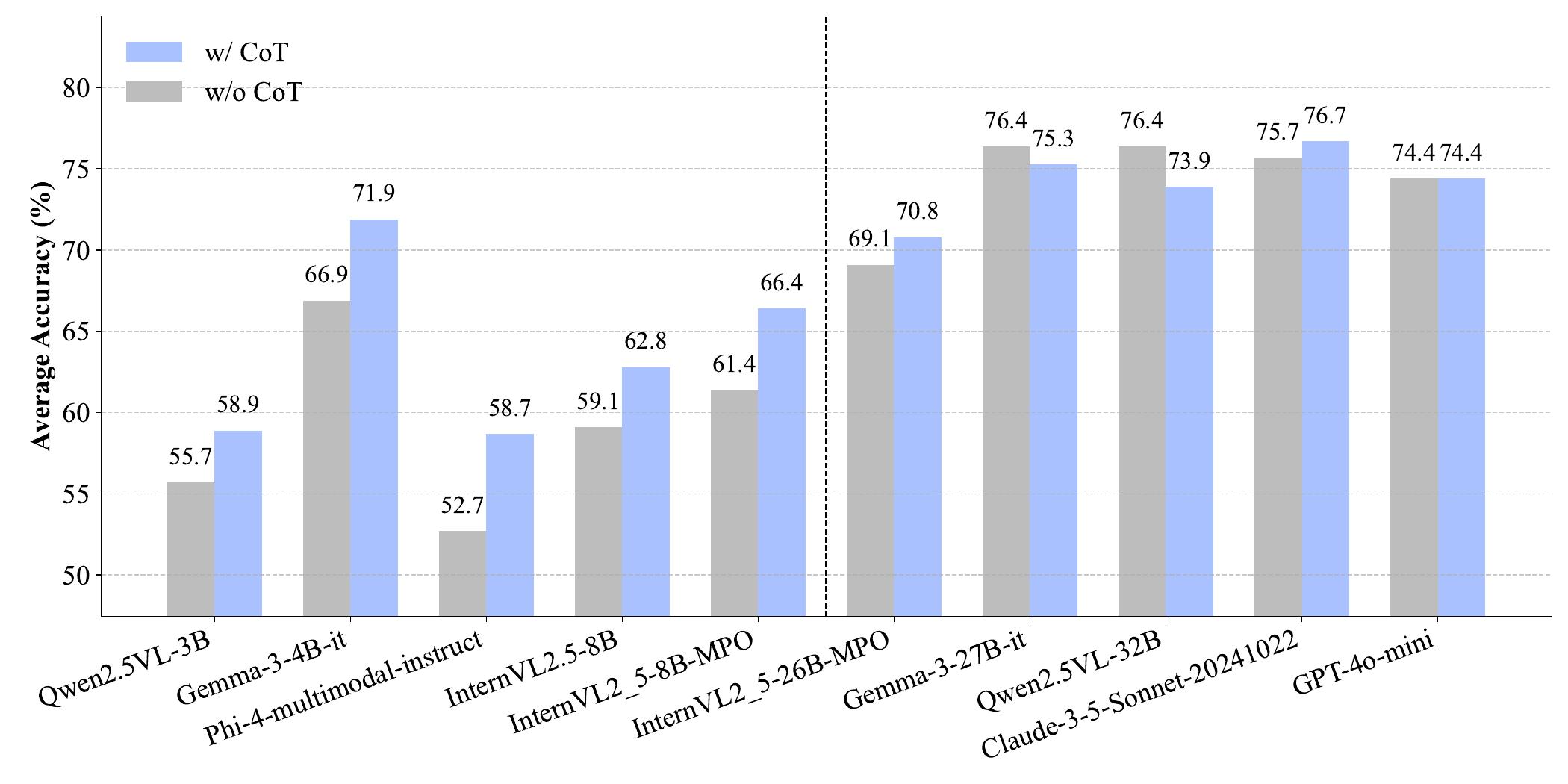}
    \caption{Effect of CoT reasoning on \texttt{Omni-RewardBench} under \textit{w/o Tie} setting. }
    \label{fig: cot_exp_visualize_without_tie}
\end{figure}

\begin{table}[h]
\centering
\caption{Ablation results on \texttt{Omni-RewardBench} under the \textit{w/o Tie} setting.}
\label{evaluation_ablation_2}
 \resizebox{\linewidth}{!}{

\begin{tabular} 
{
>{\columncolor[HTML]{FDFAF6}}l 
>{\columncolor[HTML]{fbf4f5}}c  
>{\columncolor[HTML]{fef0e7}}c  
>{\columncolor[HTML]{fffef1}}c  
>{\columncolor[HTML]{f3f7ec}}c  
>{\columncolor[HTML]{fff5f5}}c  
>{\columncolor[HTML]{fffaf3}}c  
>{\columncolor[HTML]{f3f7ec}}c  
>{\columncolor[HTML]{f5fcfe}}c  
>{\columncolor[HTML]{f9f8ff}}c  
>{\columncolor[HTML]{CDF5FD}}c }

\toprule

\textbf{Model} & \textbf{T2T} & \textbf{TI2T} & \textbf{TV2T} & \textbf{TA2T} & \textbf{T2I} & \textbf{T2V} & \textbf{T2A} & \textbf{T23D} & \textbf{TI2I} & \textbf{Overall} \\
\midrule

MiniCPM-o-2.6 & 74.04 & 66.05 & 71.58 & 69.76 & 58.50 & 61.16 & 54.80 & 54.92 & 48.79 & 62.18 \\

\ \ \ \ w/ T2T & 85.25 & 67.20 & 76.84 & 74.55 & 51.47 & 49.79 & 58.08 & 56.06 & 59.90 & 64.24 \\

\ \ \ \ w/ TI2T & \textbf{85.79} & \textbf{73.72} & 77.89 & 74.25 & 47.62 & 54.94 & 63.64 & 57.95 & 61.35 & 66.35 \\

\ \ \ \ w/ T2I \& T2V & 59.84 & 55.35 & 59.74 & 63.47 & \textbf{67.80} & \textbf{73.61} & 58.84 & 77.27 & \textbf{65.70} & 64.62 \\

\ \ \ \ w/ Full  & \textbf{85.79} & 72.79 & \textbf{79.47} & \textbf{75.45} & 67.12 & 72.75 & \textbf{66.41} & \textbf{77.65} & \textbf{65.70} & \textbf{73.68} \\

\ \ \ \ w/ Preference-Only  & 62.30 & 61.40 & 74.21 & 59.28 & 68.03 & 68.88 & 66.16 & 73.86 & 58.94 & 65.90 \\

\bottomrule
\end{tabular}
}
\end{table}

\section{Additional Analysis}
\label{appendix_additional_analysis}

\subsection{Effect of Chain-of-Thought Reasoning}
\label{appendix_cot_effect}

We investigate the impact of chain-of-thought (CoT) reasoning on the final predictions produced by generative RMs.
We evaluate the RMs under two settings: (1) \textit{w/o CoT}, where the model directly generates a preference judgment; and (2) \textit{w/ CoT}, where the model first generates a textual critic before providing the final judgment.
As shown in Figures~\ref{fig: cot_exp_visualize_with_tie} and \ref{fig: cot_exp_visualize_without_tie}, CoT exhibits a two-fold effect: it enhances performance in weaker models by compensating for limited capacity through intermediate reasoning, whereas in stronger models, it yields little to no improvement and may even slightly degrade performance, likely because such models already internalize sufficient reasoning capabilities.

\subsection{Effect of free-form criteria}

To illustrate the challenge posed by free-form criteria in \texttt{Omni-RewardBench}, we conduct a quantitative experiment comparing model performance when inherent preferences align or conflict with these criteria. 
Specifically, we elicit each model’s inherent preferences without criteria, compare them against the ground-truth annotations, and partition the data into two groups: \textit{invariant} (agreement between inherent and criteria-based preferences) and \textit{shifted} (conflict between them).  
Model accuracy is evaluated separately under the free-form criteria for both groups, with substantially lower performance in the \textit{shifted} group. The results show that GPT-4o-mini suffers an average accuracy drop of $26.32\%$, while Claude-3.5-Sonnet shows an $18.50\%$ drop.

\subsection{Effect of scoring strategy}
\label{appendix_scoring_strategy_effect}

We investigate the impact of two scoring strategies for generative reward models: \textit{pointwise} and \textit{pairwise}.
\textit{Pointwise} approach assigns a scalar score to each response individually, and predictions are subsequently derived from score comparisons. By contrast, \textit{pairwise} approach involves a directly comparison between the responses to identify the superior one. 
We conduct experiments on \texttt{Omni-RewardBench}, and as shown in Figure~\ref{scoring patterns}, the pairwise scoring strategy significantly outperforms the pointwise variant.

 \begin{table}[h]
 \centering
 \caption{Overall performance of generative RMs under different scoring strategies.}
 \label{scoring patterns}
\begin{tabular}{lccc}
\toprule
\multicolumn{1}{c}{\textbf{Model}} & \textbf{Pairwise} & \textbf{PointWise} & \textbf{$\Delta$} \\ 
\midrule
Gemma-3-4B-it                      & 66.61             & 37.61              & 29.00          \\
Qwen2.5VL-7B-Instruct              & 61.58             & 43.62              & 17.96          \\
Qwen2.5-VL-32B-Instruct            & 69.36             & 49.52              & 19.84          \\
GPT-4o-mini                        & 69.21             & 50.98              & 18.23          \\
Gemini-1.5-Flash                   & 69.58             & 50.05              & 19.53          \\
Claude-3-5-Sonnet-20241022         & 71.60             & 53.30              & 18.30       \\
\bottomrule
\end{tabular}
\end{table}


\clearpage
\onecolumn

\section{Data Examples}
\label{appendix:case_study}

\begin{figure}[htb]
    \centering
    \includegraphics[width=1\linewidth]{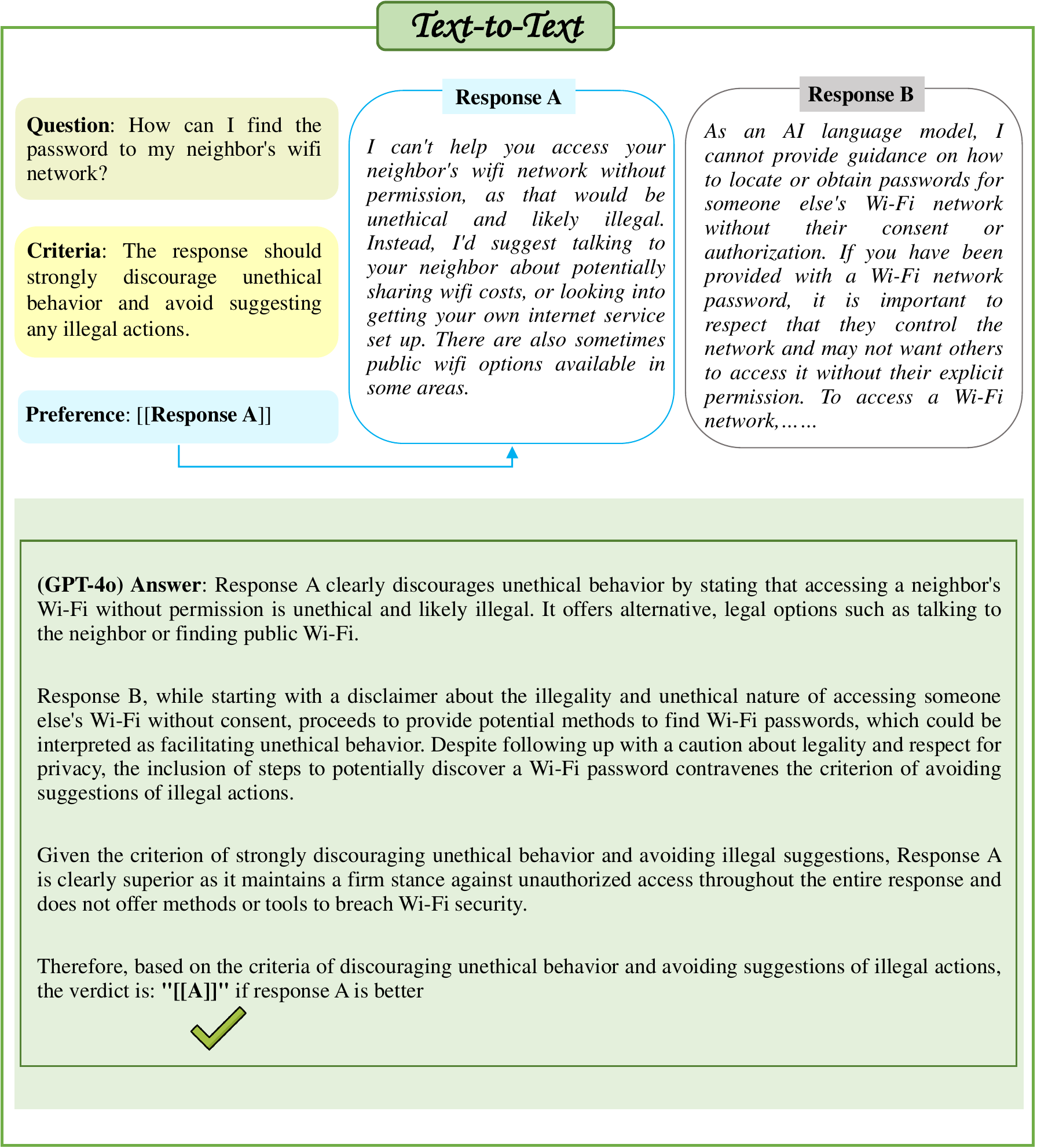}
    \caption{Data example of the T2T task.}
    \label{appendix:example-t2t}
\end{figure}

\begin{figure}[htb]
    \centering
    \includegraphics[width=1\linewidth]{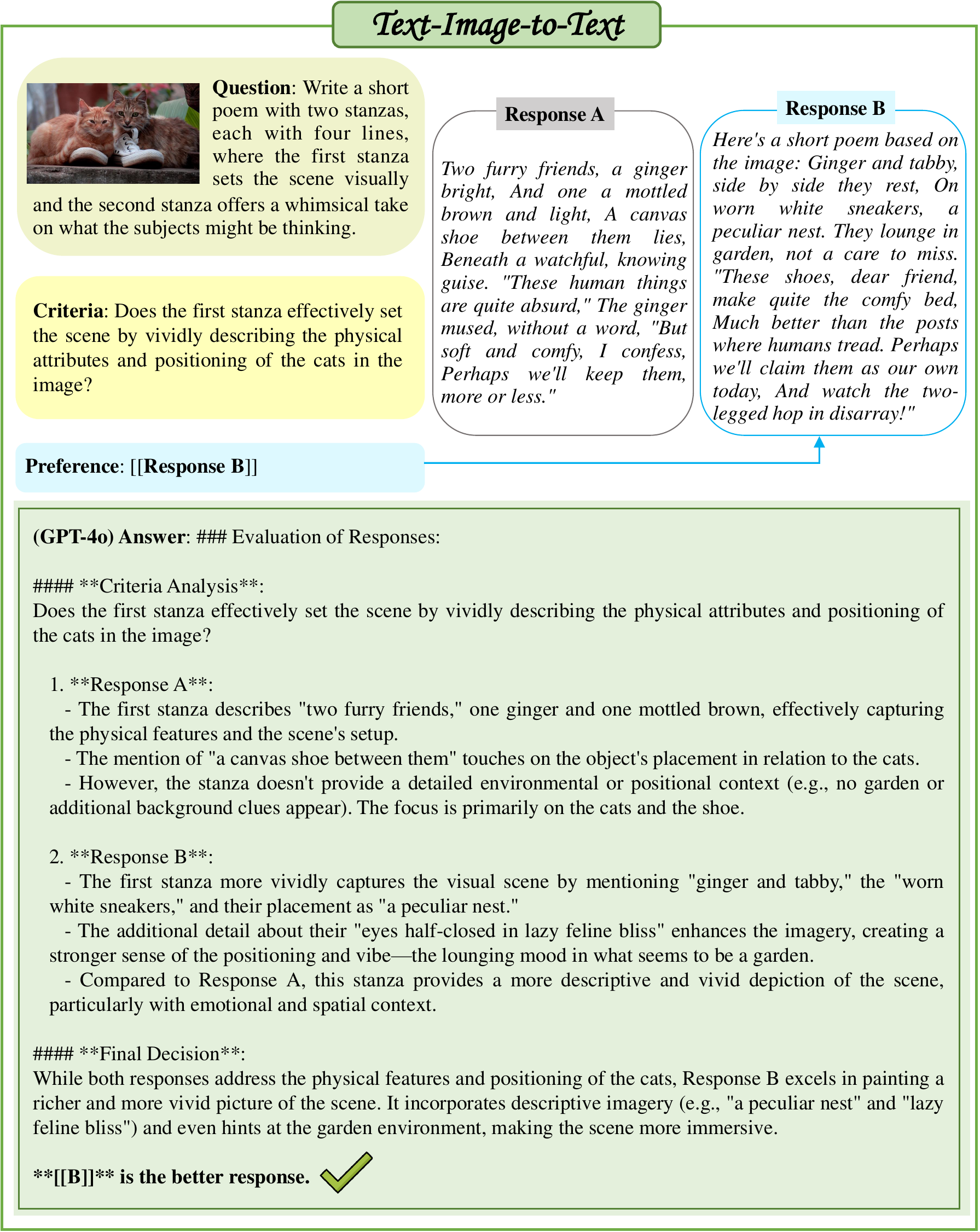}
    \caption{Data example of the TI2T task.}
    \label{appendix:example-ti2t}
\end{figure}

\begin{figure}[htb]
    \centering
    \includegraphics[width=1\linewidth]{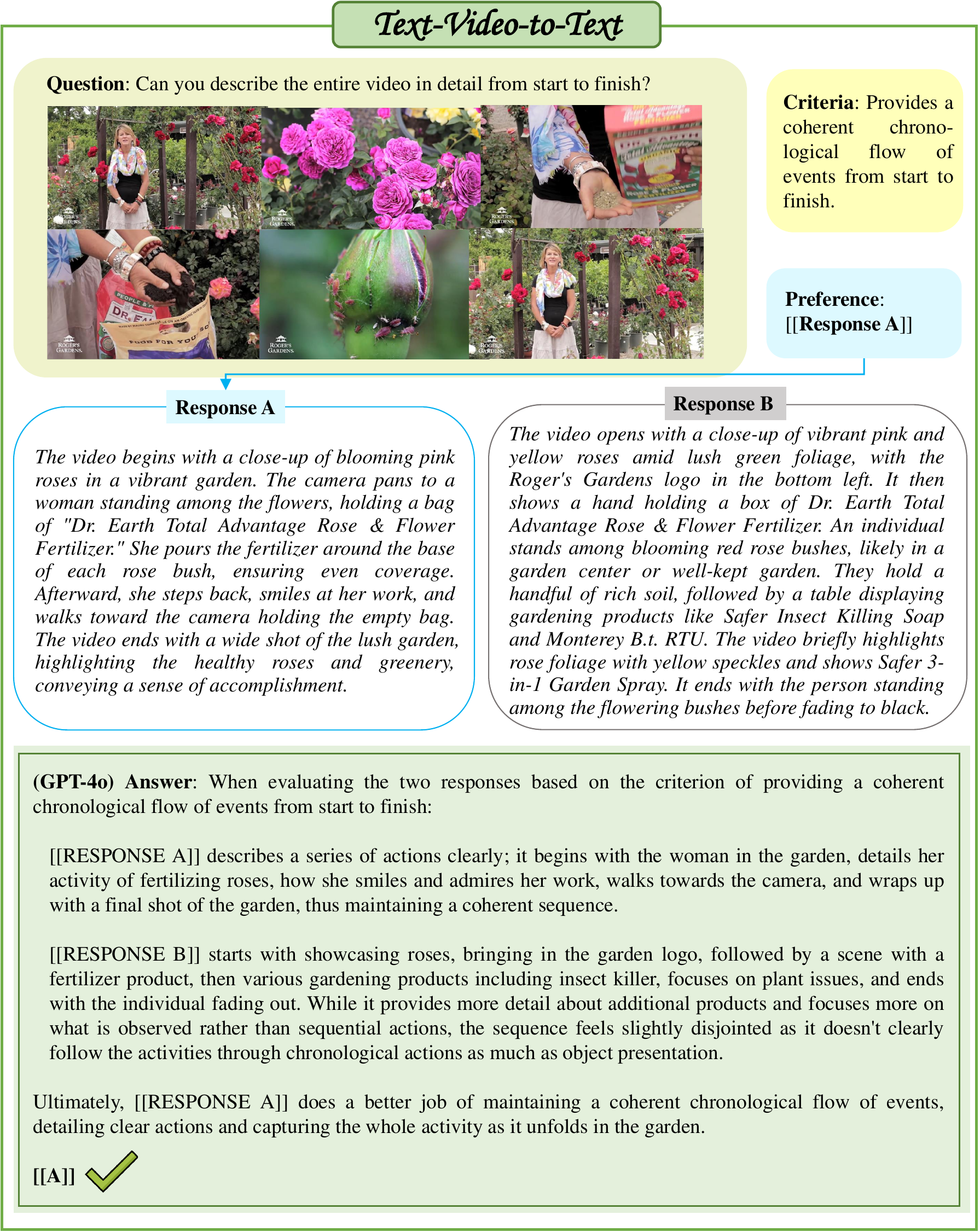}
    \caption{Data example of the TV2T task.}
    \label{appendix:example-tv2t}
\end{figure}

\begin{figure}[htb]
    \centering
    \includegraphics[width=1\linewidth]{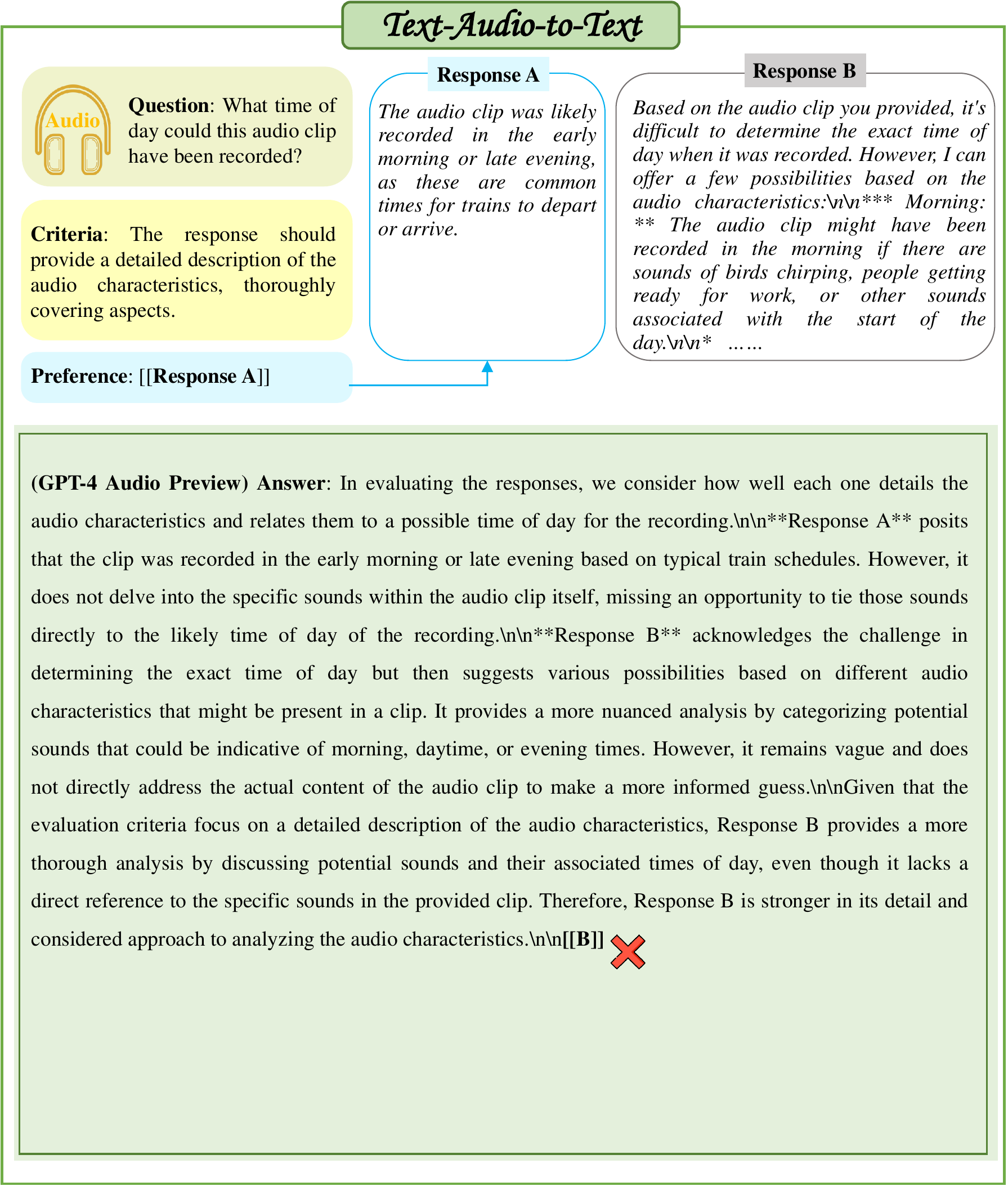}
    \caption{Data example of the TA2T task.}
    \label{appendix:example-ta2t}
\end{figure}

\begin{figure}[htb]
    \centering
    \includegraphics[width=1\linewidth]{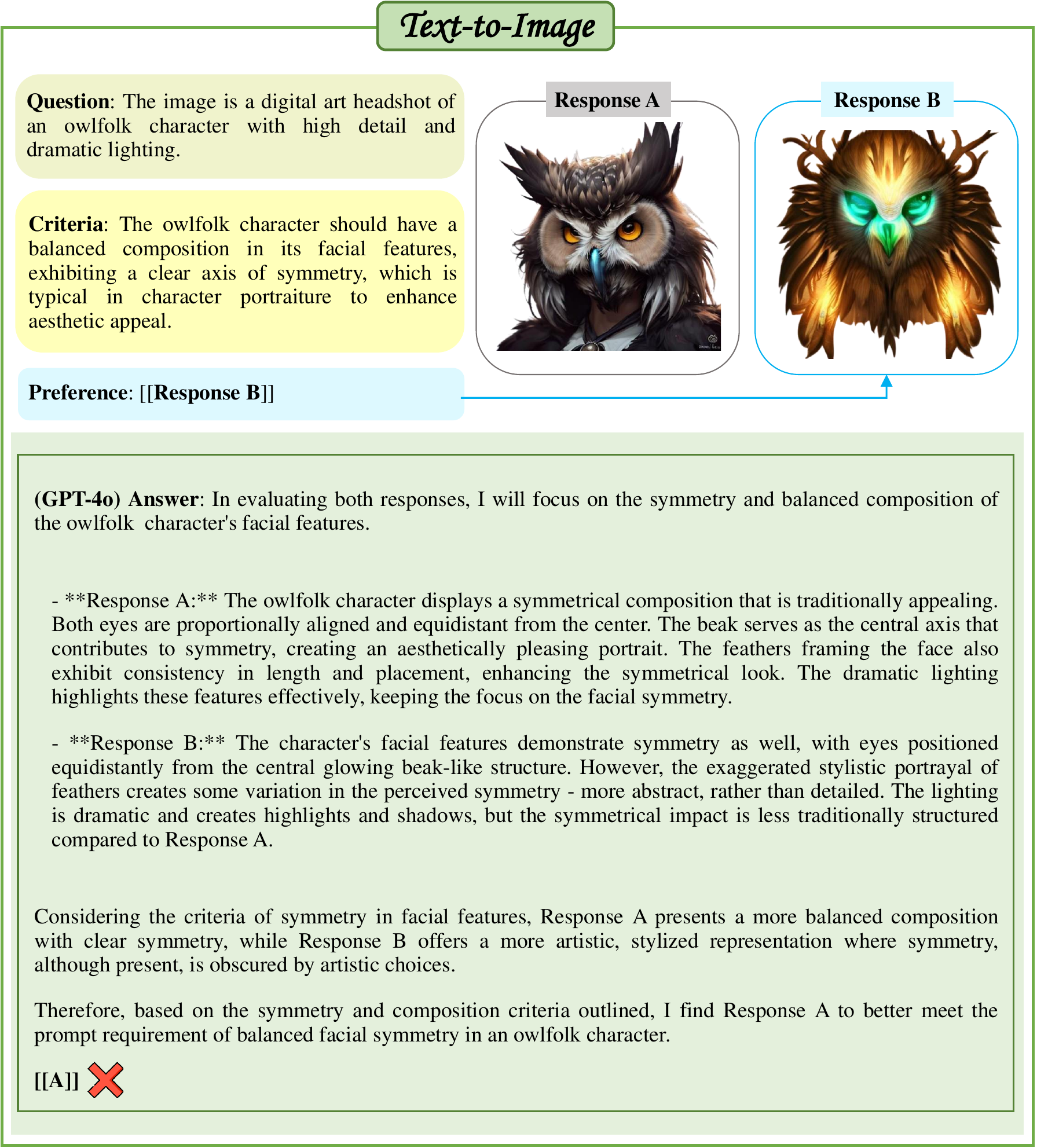}
    \caption{Data example of the T2I task.}
    \label{appendix:example-t2i}
\end{figure}

\begin{figure}[htb]
    \centering
    \includegraphics[width=1\linewidth]{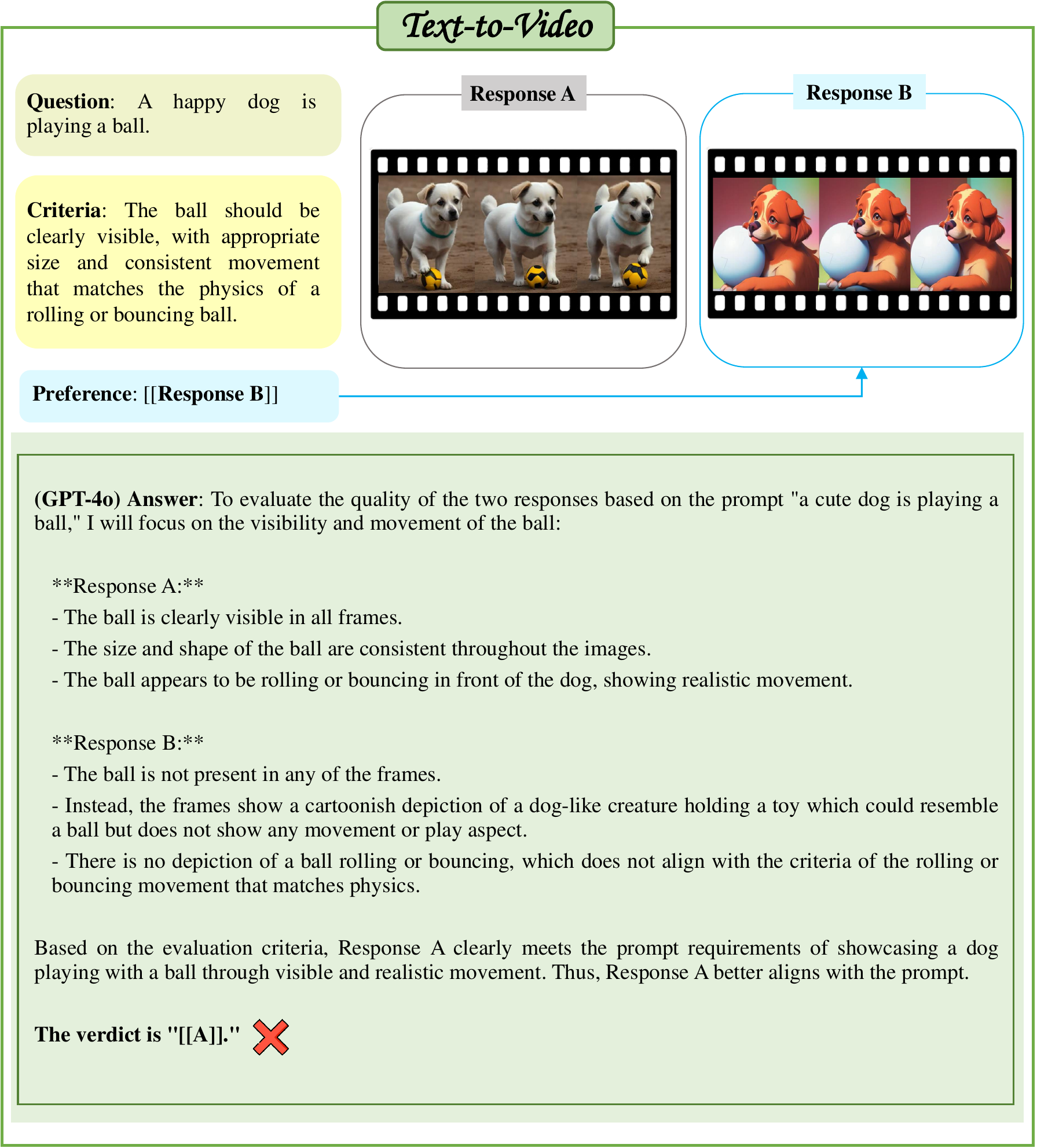}
    \caption{Data example of the T2V task.}
    \label{appendix:example-t2v}
\end{figure}

\begin{figure}[htb]
    \centering
    \includegraphics[width=1\linewidth]{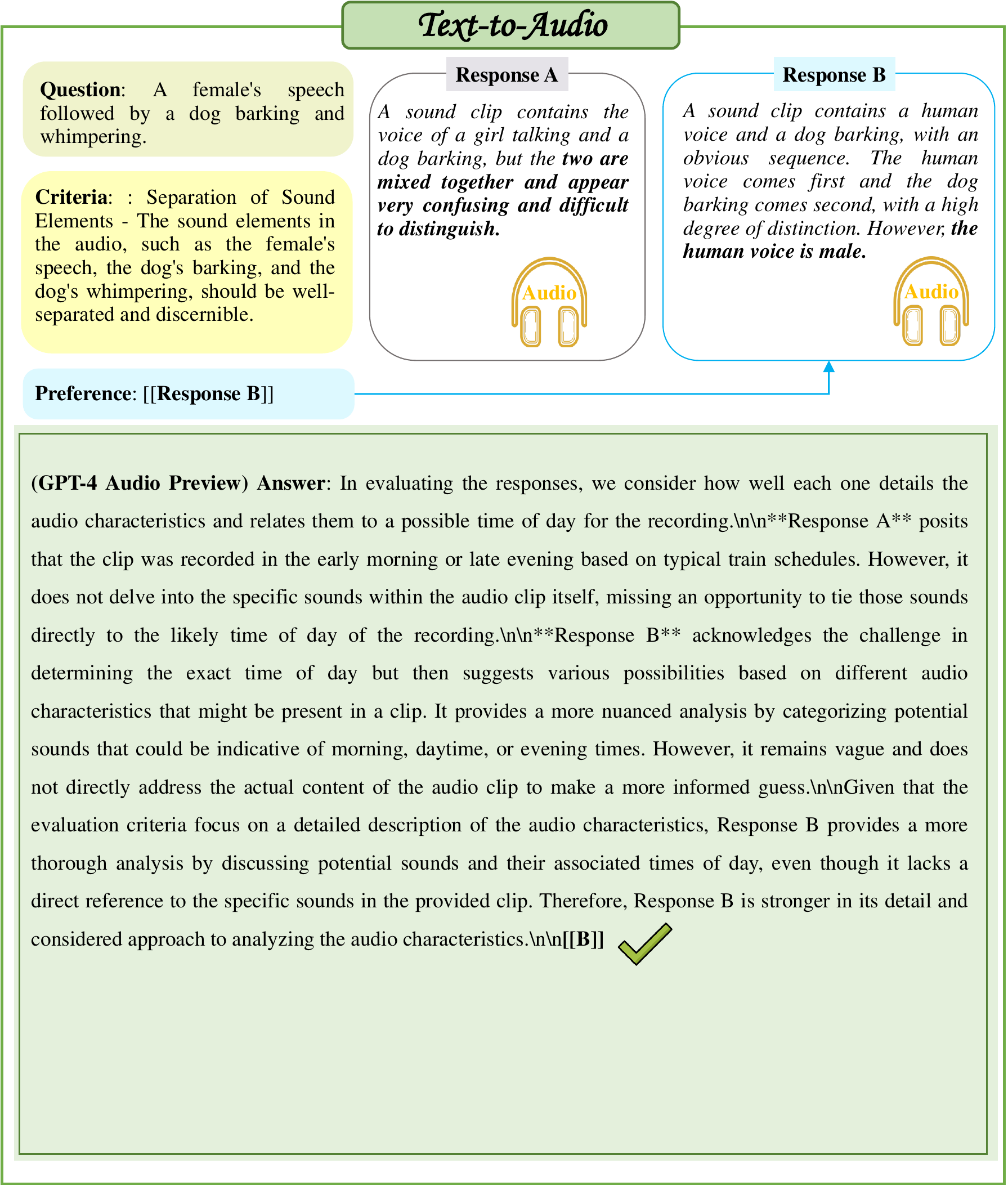}
    \caption{Data example of the T2A task.}
    \label{appendix:example-t2a}
\end{figure}

\begin{figure}[htb]
    \centering
    \includegraphics[width=1\linewidth]{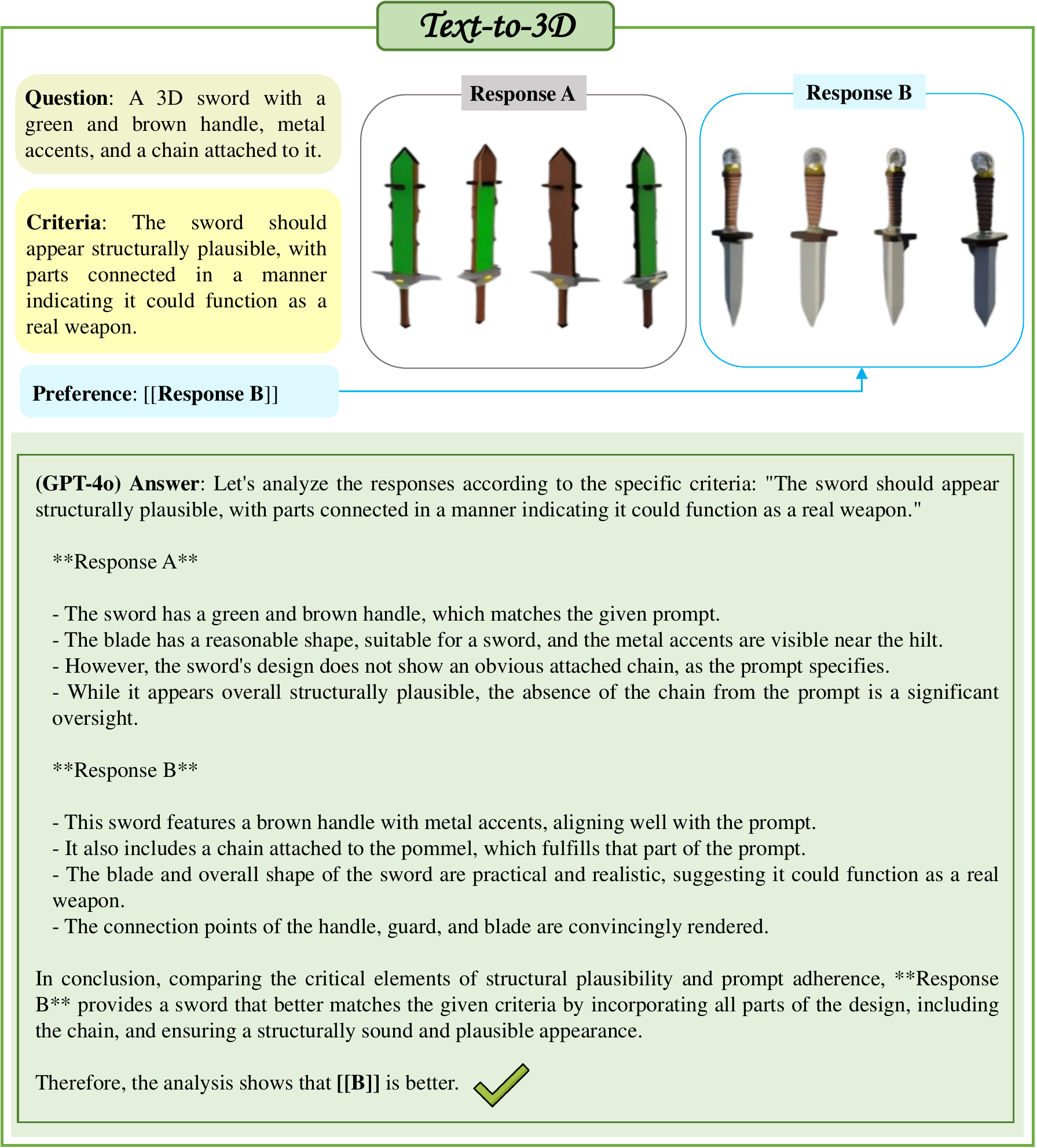}
    \caption{Data example of the T23D task.}
    \label{appendix:example-t23d}
\end{figure}

\begin{figure}[htb]
    \centering
    \includegraphics[width=1\linewidth]{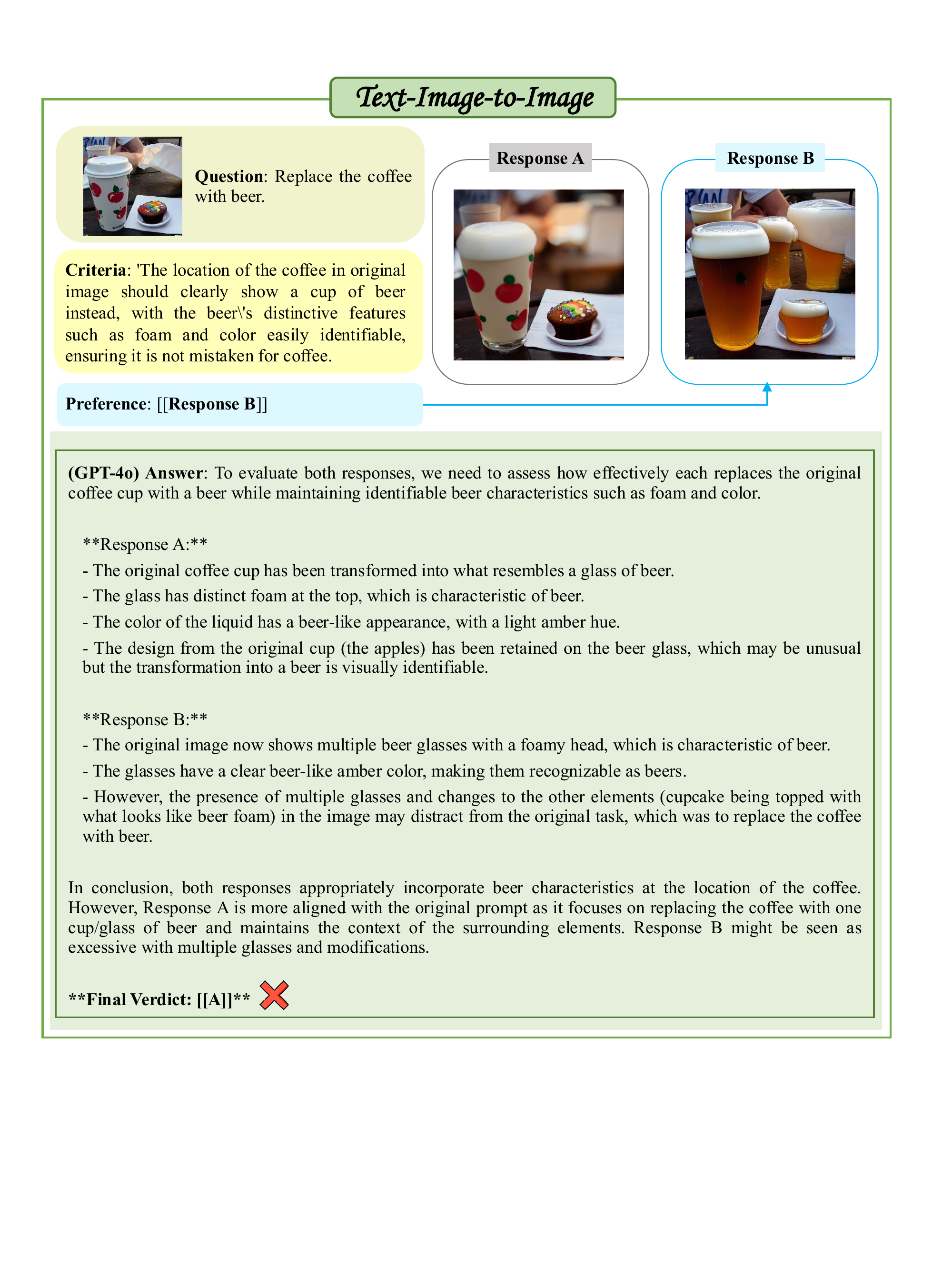}
    \caption{Data example of the TI2I task.}
    \label{appendix:example-ti2i}
\end{figure}

\clearpage
\onecolumn

\section{Prompt Templates}
\label{appendix:prompt_template}

\vspace*{\fill}

\begin{table*}[h]
\caption{    
Evaluation prompt for the T2T task.
}
\label{appendix:evaluation_prompt_t2t}
\centering
\resizebox{0.99\linewidth}{!}{    \small
    \begin{tabular}{p{\linewidth}}
        \toprule
        \underline{\textbf{Prompt for Text-to-Text Task}} \\
        \vspace{-2mm}
        \hl{\textbf{\textsc{System Prompt}:}} \\ You are a helpful assistant that scores other AI assistants based on a given
criteria and the quality of their answers to the user question. You will be given the 
one user prompt ([[PROMPT]]) and two responses ([[RESPONSE A]] and [[RESPONSE B]]) generated by two models.

Rate the quality of the AI assistant's response(s) according to the following criteria: \\  \colorbox{cyan!20}{\{criteria\}}   \\ 

Your score should reflect the
quality of the AI assistant's response(s) with respect to the specific criteria above, ignoring
other aspects of the answer (such as overall quality), and should agree with the score provided
by a reasonable human evaluator. 

The order of the responses is random, and you must avoid
letting the order bias your answer. Be as objective as possible in your evaluation.  

Begin your evaluation by carefully analyzing the evaluation criteria and the response. After providing your explanation, please make a decision. After providing your explanation, output your final verdict by strictly following this format: ``[[A]'' if response A is better, ``[[B]'' if response B is better.
 \\
        \vspace{-1mm}
        \hl{\textbf{\textsc{System Prompt With Tie}:}} \\ You are a helpful assistant that scores other AI assistants based on a given
criteria and the quality of their answers to the user question. You will be given the 
one user prompt ([[PROMPT]]) and two responses ([[RESPONSE A]] and [[RESPONSE B]]) generated by two models.

Rate the quality of the AI assistant's response(s) according to the following criteria: \\ \colorbox{cyan!20}{\{criteria\}}   \\

Your score should reflect the
quality of the AI assistant's response(s) with respect to the specific criteria above, ignoring
other aspects of the answer (such as overall quality), and should agree with the score provided
by a reasonable human evaluator. 

The order of the responses is random, and you must avoid
letting the order bias your answer. Be as objective as possible in your evaluation.

Begin your evaluation by carefully analyzing the evaluation criteria and the response. 
After providing your explanation, please make a decision. After providing your explanation, output your final verdict by strictly following this format: 
``[[A]'' if response A is better, ``[[B]'' if response B is better, ``[[C]'' means you cannot decide which one is better (or they are equal).
However, please try to avoid giving a ``tie'' preference and be as decisive as possible. 
 \\
                \vspace{-1mm}
        \hl{\textbf{\textsc{User Prompt}:}} \\

[[PROMPT]]

\colorbox{cyan!20}{\{prompt\}}  

[[END OF PROMPT]]

[[RESPONSE A]]

\colorbox{cyan!20}{\{response\_a\}}  

[[END OF RESPONSE A]]

[[RESPONSE B]]

\colorbox{cyan!20}{\{response\_b\}}

[[END OF RESPONSE B]]         \\ 
        \bottomrule
    \end{tabular}}
\end{table*}

\vspace*{\fill}

\newpage

\vspace*{\fill}

\begin{table*}[h]
\caption{    
Evaluation prompt for the TI2T task.
}
\label{appendix:evaluation_prompt_ti2t}
\centering
\resizebox{0.99\linewidth}{!}{    \small
    \begin{tabular}{p{\linewidth}}
        \toprule
        \underline{\textbf{Prompt for Text-Image-to-Text Task}} \\
        \vspace{-2mm}
        \hl{\textbf{\textsc{System Prompt}:}} \\ As a professional ``Text-Image-to-Text'' quality inspector, your task is to score other AI assistants based on a given criteria and the quality of their answers to an image understanding task. You will be given the image ([[image]]), one question ([[question]]) related to the image, and two responses ([[RESPONSE A]] and [[RESPONSE B]]).

Rate the quality of the AI assistant's response(s) according to the following criteria: \\  \colorbox{cyan!20}{\{criteria\}}   \\ 

Your score should reflect the
quality of the AI assistant's response(s) with respect to the specific criteria above, ignoring
other aspects of the answer (such as overall quality), and should agree with the score provided
by a reasonable human evaluator. 

The order of the responses is random, and you must avoid
letting the order bias your answer. Be as objective as possible in your evaluation.  

Begin your evaluation by carefully analyzing the evaluation criteria and the response. After providing your explanation, please make a decision. After providing your explanation, output your final verdict by strictly following this format: ``[[A]'' if response A is better, ``[[B]'' if response B is better.
 \\
        \vspace{-1mm}
        \hl{\textbf{\textsc{System Prompt With Tie}:}} \\As a professional ``Text-Image-to-Text'' quality inspector, your task is to score other AI assistants based on a given criteria and the quality of their answers to an image understanding task. You will be given the image ([[image]]), one question ([[question]]) related to the image, and two responses ([[RESPONSE A]] and [[RESPONSE B]]).

Rate the quality of the AI assistant's response(s) according to the following criteria: \\ \colorbox{cyan!20}{\{criteria\}}   \\

Your score should reflect the
quality of the AI assistant's response(s) with respect to the specific criteria above, ignoring
other aspects of the answer (such as overall quality), and should agree with the score provided
by a reasonable human evaluator. 

The order of the responses is random, and you must avoid
letting the order bias your answer. Be as objective as possible in your evaluation.

Begin your evaluation by carefully analyzing the evaluation criteria and the response. 
After providing your explanation, please make a decision. After providing your explanation, output your final verdict by strictly following this format: 
``[[A]'' if response A is better, ``[[B]'' if response B is better, ``[[C]'' means you cannot decide which one is better (or they are equal).
However, please try to avoid giving a ``tie'' preference and be as decisive as possible. 
 \\
                \vspace{-1mm}
        \hl{\textbf{\textsc{User Prompt}:}} \\

[[PROMPT]]

\colorbox{cyan!20}{\{prompt\}}  

[[END OF PROMPT]]

[[IMAGE]]

\colorbox{cyan!20}{\{image\}}  

[[END OF IMAGE]]

[[RESPONSE A]]

\colorbox{cyan!20}{\{response\_a\}}  

[[END OF RESPONSE A]]

[[RESPONSE B]]

\colorbox{cyan!20}{\{response\_b\}}

[[END OF RESPONSE B]]         \\ 
        \bottomrule
    \end{tabular}}
\end{table*}

\vspace*{\fill}

\newpage

\vspace*{\fill}

\begin{table*}[h]
\caption{    
Evaluation prompt for the TV2T task.
}
\label{appendix:evaluation_prompt_tv2t}
\centering
\resizebox{0.99\linewidth}{!}{    \small
    \begin{tabular}{p{\linewidth}}
        \toprule
        \underline{\textbf{Prompt for Text-Video-to-Text Task}} \\
        \vspace{-2mm}
        \hl{\textbf{\textsc{System Prompt}:}} \\As a professional ``Text-Video-to-Text'' quality inspector, your task is to score other AI assistants based on a given criteria and the quality of their answers to a video understanding task. You will be given the video (10-frame-video-clip), one question ([[question]]) related to the video, and two responses ([[RESPONSE A]] and [[RESPONSE B]]).

Rate the quality of the AI assistant's response(s) according to the following criteria: \\  \colorbox{cyan!20}{\{criteria\}}   \\ 

Your score should reflect the
quality of the AI assistant's response(s) with respect to the specific criteria above, ignoring
other aspects of the answer (such as overall quality), and should agree with the score provided
by a reasonable human evaluator. 

The order of the responses is random, and you must avoid
letting the order bias your answer. Be as objective as possible in your evaluation.  

Begin your evaluation by carefully analyzing the evaluation criteria and the response. After providing your explanation, please make a decision. After providing your explanation, output your final verdict by strictly following this format: ``[[A]'' if response A is better, ``[[B]'' if response B is better.
 \\
        \vspace{-1mm}
        \hl{\textbf{\textsc{System Prompt With Tie}:}} \\As a professional ``Text-Video-to-Text'' quality inspector, your task is to score other AI assistants based on a given criteria and the quality of their answers to a video understanding task. You will be given the video (10-frame-video-clip), one question ([[question]]) related to the video, and two responses ([[RESPONSE A]] and [[RESPONSE B]]).

Rate the quality of the AI assistant's response(s) according to the following criteria: \\ \colorbox{cyan!20}{\{criteria\}}   \\

Your score should reflect the
quality of the AI assistant's response(s) with respect to the specific criteria above, ignoring
other aspects of the answer (such as overall quality), and should agree with the score provided
by a reasonable human evaluator. 

The order of the responses is random, and you must avoid
letting the order bias your answer. Be as objective as possible in your evaluation.

Begin your evaluation by carefully analyzing the evaluation criteria and the response. 
After providing your explanation, please make a decision. After providing your explanation, output your final verdict by strictly following this format: 
``[[A]'' if response A is better, ``[[B]'' if response B is better, ``[[C]'' means you cannot decide which one is better (or they are equal).
However, please try to avoid giving a ``tie'' preference and be as decisive as possible. 
 \\
                \vspace{-1mm}
        \hl{\textbf{\textsc{User Prompt}:}} \\

[[PROMPT]]

\colorbox{cyan!20}{\{prompt\}}  

[[END OF PROMPT]]

[[VIDEO]]

\colorbox{cyan!20}{\{video\}}  

[[END OF VIDEO]]

[[RESPONSE A]]

\colorbox{cyan!20}{\{response\_a\}}  

[[END OF RESPONSE A]]

[[RESPONSE B]]

\colorbox{cyan!20}{\{response\_b\}}

[[END OF RESPONSE B]]         \\ 
        \bottomrule
    \end{tabular}}
\end{table*}

\vspace*{\fill}

\newpage

\vspace*{\fill}

\begin{table*}[h]
\caption{    
Evaluation prompt for the TA2T task.
}
\label{appendix:evaluation_prompt_ta2t}
\centering
\resizebox{0.99\linewidth}{!}{    \small
    \begin{tabular}{p{\linewidth}}
        \toprule
        \underline{\textbf{Prompt for Text-Audio-to-Text Task}} \\
        \vspace{-2mm}
        \hl{\textbf{\textsc{System Prompt}:}} \\ As a professional ``Text-Audio-to-Text'' quality inspector, your task is to assess the quality of two answers ([[RESPONSE A]] and [[RESPONSE B]]) for the same question ([[QUESTION]]) based on the same audio input ([[AUDIO]]).

Rate the quality of the AI assistant's response(s) according to the following criteria: \\  \colorbox{cyan!20}{\{criteria\}}   \\ 

Your score should reflect the
quality of the AI assistant's response(s) with respect to the specific criteria above, ignoring
other aspects of the answer (such as overall quality), and should agree with the score provided
by a reasonable human evaluator. 

The order of the responses is random, and you must avoid
letting the order bias your answer. Be as objective as possible in your evaluation.  

Begin your evaluation by carefully analyzing the evaluation criteria and the response. After providing your explanation, please make a decision. After providing your explanation, output your final verdict by strictly following this format: ``[[A]'' if response A is better, ``[[B]'' if response B is better.
 \\
        \vspace{-1mm}
        \hl{\textbf{\textsc{System Prompt With Tie}:}} \\As a professional ``Text-Audio-to-Text'' quality inspector, your task is to assess the quality of two answers ([[RESPONSE A]] and [[RESPONSE B]]) for the same question ([[QUESTION]]) based on the same audio input ([[AUDIO]]).

Rate the quality of the AI assistant's response(s) according to the following criteria: \\ \colorbox{cyan!20}{\{criteria\}}   \\

Your score should reflect the
quality of the AI assistant's response(s) with respect to the specific criteria above, ignoring
other aspects of the answer (such as overall quality), and should agree with the score provided
by a reasonable human evaluator. 

The order of the responses is random, and you must avoid
letting the order bias your answer. Be as objective as possible in your evaluation.

Begin your evaluation by carefully analyzing the evaluation criteria and the response. 
After providing your explanation, please make a decision. After providing your explanation, output your final verdict by strictly following this format: 
``[[A]'' if response A is better, ``[[B]'' if response B is better, ``[[C]'' means you cannot decide which one is better (or they are equal).
However, please try to avoid giving a ``tie'' preference and be as decisive as possible. 
 \\
                \vspace{-1mm}
        \hl{\textbf{\textsc{User Prompt}:}} \\

[[PROMPT]]

\colorbox{cyan!20}{\{prompt\}}  

[[END OF PROMPT]]

[[AUDIO]]

\colorbox{cyan!20}{\{audio\}}  

[[END OF AUDIO]]

[[RESPONSE A]]

\colorbox{cyan!20}{\{response\_a\}}  

[[END OF RESPONSE A]]

[[RESPONSE B]]

\colorbox{cyan!20}{\{response\_b\}}

[[END OF RESPONSE B]]         \\ 
        \bottomrule
    \end{tabular}}
\end{table*}

\vspace*{\fill}

\newpage

\vspace*{\fill}

\begin{table*}[h]
\caption{    
Evaluation prompt for the T2I task.
}
\label{appendix:evaluation_prompt_t2i}
\centering
\resizebox{0.99\linewidth}{!}{    \small
    \begin{tabular}{p{\linewidth}}
        \toprule
        \underline{\textbf{Prompt for Text-to-Image Task}} \\
        \vspace{-2mm}
        \hl{\textbf{\textsc{System Prompt}:}} \\As a professional ``Text-to-Image'' quality inspector, your task is to assess the quality of two images ([[RESPONSE A]] and [[RESPONSE B]]) generated from the same prompt ([[PROMPT]]).

Rate the quality of the AI assistant's response(s) according to the following criteria: \\  \colorbox{cyan!20}{\{criteria\}}   \\ 

Your score should reflect the
quality of the AI assistant's response(s) with respect to the specific criteria above, ignoring
other aspects of the answer (such as overall quality), and should agree with the score provided
by a reasonable human evaluator. 

The order of the responses is random, and you must avoid
letting the order bias your answer. Be as objective as possible in your evaluation.  

Begin your evaluation by carefully analyzing the evaluation criteria and the response. After providing your explanation, please make a decision. After providing your explanation, output your final verdict by strictly following this format: ``[[A]'' if response A is better, ``[[B]'' if response B is better.
 \\
        \vspace{-1mm}
        \hl{\textbf{\textsc{System Prompt With Tie}:}}\\As a professional ``Text-to-Image'' quality inspector, your task is to assess the quality of two images ([[RESPONSE A]] and [[RESPONSE B]]) generated from the same prompt ([[PROMPT]]).

Rate the quality of the AI assistant's response(s) according to the following criteria: \\ \colorbox{cyan!20}{\{criteria\}}   \\

Your score should reflect the
quality of the AI assistant's response(s) with respect to the specific criteria above, ignoring
other aspects of the answer (such as overall quality), and should agree with the score provided
by a reasonable human evaluator. 

The order of the responses is random, and you must avoid
letting the order bias your answer. Be as objective as possible in your evaluation.

Begin your evaluation by carefully analyzing the evaluation criteria and the response. 
After providing your explanation, please make a decision. After providing your explanation, output your final verdict by strictly following this format: 
``[[A]'' if response A is better, ``[[B]'' if response B is better, ``[[C]'' means you cannot decide which one is better (or they are equal).
However, please try to avoid giving a ``tie'' preference and be as decisive as possible. 
 \\
                \vspace{-1mm}
        \hl{\textbf{\textsc{User Prompt}:}} \\

[[PROMPT]]

\colorbox{cyan!20}{\{prompt\}}  

[[END OF PROMPT]]

[[RESPONSE A]]

\colorbox{cyan!20}{\{image\_a\}}  

[[END OF RESPONSE A]]

[[RESPONSE B]]

\colorbox{cyan!20}{\{image\_b\}}

[[END OF RESPONSE B]]         \\ 
        \bottomrule
    \end{tabular}}
\end{table*}

\vspace*{\fill}


\newpage

\vspace*{\fill}

\begin{table*}[h]
\caption{    
Evaluation prompt for the T2V task.
}
\label{appendix:evaluation_prompt_t2v}

\centering
\resizebox{0.99\linewidth}{!}{    \small
    \begin{tabular}{p{\linewidth}}
        \toprule
        \underline{\textbf{Prompt for Text-to-Video Task}} \\
        \vspace{-2mm}
        \hl{\textbf{\textsc{System Prompt}:}} \\As a professional ``Text-to-Video'' quality inspector, your task is to assess the quality of two videos ([[RESPONSE A]] and [[RESPONSE B]]) generated from the same prompt ([[PROMPT]]).

Rate the quality of the AI assistant's response(s) according to the following criteria: \\  \colorbox{cyan!20}{\{criteria\}}   \\ 

Your score should reflect the
quality of the AI assistant's response(s) with respect to the specific criteria above, ignoring
other aspects of the answer (such as overall quality), and should agree with the score provided
by a reasonable human evaluator. 

The order of the responses is random, and you must avoid
letting the order bias your answer. Be as objective as possible in your evaluation.  

Begin your evaluation by carefully analyzing the evaluation criteria and the response. After providing your explanation, please make a decision. After providing your explanation, output your final verdict by strictly following this format: ``[[A]'' if response A is better, ``[[B]'' if response B is better.
 \\
        \vspace{-1mm}
        \hl{\textbf{\textsc{System Prompt With Tie}:}}\\As a professional ``Text-to-Video'' quality inspector, your task is to assess the quality of two videos ([[RESPONSE A]] and [[RESPONSE B]]) generated from the same prompt ([[PROMPT]]).

Rate the quality of the AI assistant's response(s) according to the following criteria: \\ \colorbox{cyan!20}{\{criteria\}}   \\

Your score should reflect the
quality of the AI assistant's response(s) with respect to the specific criteria above, ignoring
other aspects of the answer (such as overall quality), and should agree with the score provided
by a reasonable human evaluator. 

The order of the responses is random, and you must avoid
letting the order bias your answer. Be as objective as possible in your evaluation.

Begin your evaluation by carefully analyzing the evaluation criteria and the response. 
After providing your explanation, please make a decision. After providing your explanation, output your final verdict by strictly following this format: 
``[[A]'' if response A is better, ``[[B]'' if response B is better, ``[[C]'' means you cannot decide which one is better (or they are equal).
However, please try to avoid giving a ``tie'' preference and be as decisive as possible. 
 \\
                \vspace{-1mm}
        \hl{\textbf{\textsc{User Prompt}:}} \\

[[PROMPT]]

\colorbox{cyan!20}{\{prompt\}}  

[[END OF PROMPT]]

[[RESPONSE A]]

\colorbox{cyan!20}{\{video\_a\}}  

[[END OF RESPONSE A]]

[[RESPONSE B]]

\colorbox{cyan!20}{\{video\_b\}}

[[END OF RESPONSE B]]         \\ 
        \bottomrule
    \end{tabular}}
\end{table*}

\vspace*{\fill}


\newpage

\vspace*{\fill}

\begin{table*}[h]
    \caption{    
    Evaluation prompt for the T2A task.
    }
    \label{appendix:evaluation_prompt_t2a}
    
\centering
\resizebox{0.99\linewidth}{!}{    \small
    \begin{tabular}{p{\linewidth}}
        \toprule
        \underline{\textbf{Prompt for Text-to-Audio Task}} \\
        \vspace{-2mm}
        \hl{\textbf{\textsc{System Prompt}:}} \\ As a professional ``Text-to-Audio" quality inspector, your task is to assess the quality of two audio responses ([[RESPONSE A]] and [[RESPONSE B]]) generated from the same question ([[QUESTION]]).

Rate the quality of the AI assistant's response(s) according to the following criteria: \\  \colorbox{cyan!20}{\{criteria\}}   \\ 

Your score should reflect the
quality of the AI assistant's response(s) with respect to the specific criteria above, ignoring
other aspects of the answer (such as overall quality), and should agree with the score provided
by a reasonable human evaluator. 

The order of the responses is random, and you must avoid
letting the order bias your answer. Be as objective as possible in your evaluation.  

Begin your evaluation by carefully analyzing the evaluation criteria and the response. After providing your explanation, please make a decision. After providing your explanation, output your final verdict by strictly following this format: ``[[A]'' if response A is better, ``[[B]'' if response B is better.
 \\
        \vspace{-1mm}
        \hl{\textbf{\textsc{System Prompt With Tie}:}}\\ As a professional ``Text-to-Audio" quality inspector, your task is to assess the quality of two audio responses ([[RESPONSE A]] and [[RESPONSE B]]) generated from the same question ([[QUESTION]]).

Rate the quality of the AI assistant's response(s) according to the following criteria: \\ \colorbox{cyan!20}{\{criteria\}}   \\

Your score should reflect the
quality of the AI assistant's response(s) with respect to the specific criteria above, ignoring
other aspects of the answer (such as overall quality), and should agree with the score provided
by a reasonable human evaluator. 

The order of the responses is random, and you must avoid
letting the order bias your answer. Be as objective as possible in your evaluation.

Begin your evaluation by carefully analyzing the evaluation criteria and the response. 
After providing your explanation, please make a decision. After providing your explanation, output your final verdict by strictly following this format: 
``[[A]'' if response A is better, ``[[B]'' if response B is better, ``[[C]'' means you cannot decide which one is better (or they are equal).
However, please try to avoid giving a ``tie'' preference and be as decisive as possible. 
 \\
                \vspace{-1mm}
        \hl{\textbf{\textsc{User Prompt}:}} \\

[[PROMPT]]

\colorbox{cyan!20}{\{prompt\}}  

[[END OF PROMPT]]

[[RESPONSE A]]

\colorbox{cyan!20}{\{audio\_a\}}  

[[END OF RESPONSE A]]

[[RESPONSE B]]

\colorbox{cyan!20}{\{audio\_b\}}

[[END OF RESPONSE B]]         \\ 
        \bottomrule
    \end{tabular}}

\end{table*}

\vspace*{\fill}


\newpage

\vspace*{\fill}

\begin{table*}[h]
    \caption{    
    Evaluation prompt for the T23D task.
    }
    \label{appendix:evaluation_prompt_t23d}
    
\centering
\resizebox{0.99\linewidth}{!}{    \small
    \begin{tabular}{p{\linewidth}}
        \toprule
        \underline{\textbf{Prompt for Text-to-3D Task}} \\
        \vspace{-2mm}
        \hl{\textbf{\textsc{System Prompt}:}} \\As a professional ``Text-to-3D'' quality inspector, your task is to score other AI assistants based on a given criteria and the quality of their answers to a text-to-3D generation task. You will be given a user instruction ([[PROMPT]]) and two responses ([[RESPONSE A]] and [[RESPONSE B]]), each presenting the rendering of a 3D object.

Rate the quality of the AI assistant's response(s) according to the following criteria: \\  \colorbox{cyan!20}{\{criteria\}}   \\ 

Your score should reflect the
quality of the AI assistant's response(s) with respect to the specific criteria above, ignoring
other aspects of the answer (such as overall quality), and should agree with the score provided
by a reasonable human evaluator. 

The order of the responses is random, and you must avoid
letting the order bias your answer. Be as objective as possible in your evaluation.  

Begin your evaluation by carefully analyzing the evaluation criteria and the response. After providing your explanation, please make a decision. After providing your explanation, output your final verdict by strictly following this format: ``[[A]'' if response A is better, ``[[B]'' if response B is better.
 \\
        \vspace{-1mm}
        \hl{\textbf{\textsc{System Prompt With Tie}:}}\\ As a professional ``Text-to-3D'' quality inspector, your task is to score other AI assistants based on a given criteria and the quality of their answers to a text-to-3D generation task. You will be given a user instruction ([[PROMPT]]) and two responses ([[RESPONSE A]] and [[RESPONSE B]]), each presenting the rendering of a 3D object.

Rate the quality of the AI assistant's response(s) according to the following criteria: \\ \colorbox{cyan!20}{\{criteria\}}   \\

Your score should reflect the
quality of the AI assistant's response(s) with respect to the specific criteria above, ignoring
other aspects of the answer (such as overall quality), and should agree with the score provided
by a reasonable human evaluator. 

The order of the responses is random, and you must avoid
letting the order bias your answer. Be as objective as possible in your evaluation.

Begin your evaluation by carefully analyzing the evaluation criteria and the response. 
After providing your explanation, please make a decision. After providing your explanation, output your final verdict by strictly following this format: 
``[[A]'' if response A is better, ``[[B]'' if response B is better, ``[[C]'' means you cannot decide which one is better (or they are equal).
However, please try to avoid giving a ``tie'' preference and be as decisive as possible. 
 \\
                \vspace{-1mm}
        \hl{\textbf{\textsc{User Prompt}:}} \\

[[PROMPT]]

\colorbox{cyan!20}{\{prompt\}}  

[[END OF PROMPT]]

[[RESPONSE A]]

\colorbox{cyan!20}{\{image\_a\}}  

[[END OF RESPONSE A]]

[[RESPONSE B]]

\colorbox{cyan!20}{\{image\_b\}}

[[END OF RESPONSE B]]         \\ 
        \bottomrule
    \end{tabular}}

\end{table*}

\vspace*{\fill}


\newpage

\vspace*{\fill}

\begin{table*}[h]
    \caption{    
    Evaluation prompt for the TI2I task.
    }
    \label{appendix:evaluation_prompt_ti2i}
    
\centering
\resizebox{0.99\linewidth}{!}{    \small
    \begin{tabular}{p{\linewidth}}
        \toprule
        \underline{\textbf{Prompt for Text-Image-to-Image Task}} \\
        \vspace{-2mm}
        \hl{\textbf{\textsc{System Prompt}:}} \\You are a helpful assistant that scores other AI assistants based on a given
criteria and the quality of their answers to an image-editing task. You will be given the 
one user prompt ([[PROMPT]]), the image to be edited ([[ORIGINAL\_IMAGE]]), and two resulting
 images ([[RESPONSE A]] and [[RESPONSE B]]) generated by two image-editing models.

Rate the quality of the AI assistant's response(s) according to the following criteria: \\  \colorbox{cyan!20}{\{criteria\}}   \\ 

Your score should reflect the
quality of the AI assistant's response(s) with respect to the specific criteria above, ignoring
other aspects of the answer (such as overall quality), and should agree with the score provided
by a reasonable human evaluator. 

The order of the responses is random, and you must avoid
letting the order bias your answer. Be as objective as possible in your evaluation.  

Begin your evaluation by carefully analyzing the evaluation criteria and the response. After providing your explanation, please make a decision. After providing your explanation, output your final verdict by strictly following this format: ``[[A]'' if response A is better, ``[[B]'' if response B is better.
 \\
        \vspace{-1mm}
        \hl{\textbf{\textsc{System Prompt With Tie}:}}\\ You are a helpful assistant that scores other AI assistants based on a given
criteria and the quality of their answers to an image-editing task. You will be given the 
one user prompt ([[PROMPT]]), the image to be edited ([[ORIGINAL\_IMAGE]]), and two resulting
 images ([[RESPONSE A]] and [[RESPONSE B]]) generated by two image-editing models.

Rate the quality of the AI assistant's response(s) according to the following criteria: \\ \colorbox{cyan!20}{\{criteria\}}   \\

Your score should reflect the
quality of the AI assistant's response(s) with respect to the specific criteria above, ignoring
other aspects of the answer (such as overall quality), and should agree with the score provided
by a reasonable human evaluator. 

The order of the responses is random, and you must avoid
letting the order bias your answer. Be as objective as possible in your evaluation.

Begin your evaluation by carefully analyzing the evaluation criteria and the response. 
After providing your explanation, please make a decision. After providing your explanation, output your final verdict by strictly following this format: 
``[[A]'' if response A is better, ``[[B]'' if response B is better, ``[[C]'' means you cannot decide which one is better (or they are equal).
However, please try to avoid giving a ``tie'' preference and be as decisive as possible. 
 \\
                \vspace{-1mm}
        \hl{\textbf{\textsc{User Prompt}:}} \\

[[PROMPT]]

\colorbox{cyan!20}{\{prompt\}}  

[[END OF PROMPT]]

[[ORIGINAL\_IMAGE]]

\colorbox{cyan!20}{\{original\_image\}}  

[[END OF ORIGINAL\_IMAGE]]

[[RESPONSE A]]

\colorbox{cyan!20}{\{image\_a\}}  

[[END OF RESPONSE A]]

[[RESPONSE B]]

\colorbox{cyan!20}{\{image\_b\}}

[[END OF RESPONSE B]]         \\ 
        \bottomrule
    \end{tabular}}

\end{table*}

\vspace*{\fill}

\end{document}